\documentclass{article}

\usepackage{amsmath}
\usepackage{microtype}
\usepackage{graphicx}
\usepackage{booktabs}
\usepackage{dblfloatfix}

\usepackage{color}

\PassOptionsToPackage{hyphens}{url}
\usepackage[accepted]{mlsys2025}
\usepackage{hyperref}
\hypersetup{pdftitle={BlockLock: Reproducible LLM Inference with Fixed Batch Geometryl}} 

\mlsystitlerunning{BlockLock: Reproducible LLM Inference with Fixed Batch Geometry} 

\begin{document}

\twocolumn[
\mlsystitle{BlockLock: Reproducible LLM Inference with Fixed Batch Geometry}

\begin{mlsysauthorlist}
\mlsysauthor{Xiuyuan Qi}{uestc}
\mlsysauthor{Pedro Reviriego}{upm}
\mlsysauthor{Javier Conde}{upm}
\mlsysauthor{Shanshan Liu}{uestc}
\end{mlsysauthorlist}
\mlsysaffiliation{uestc}{University of Electronic Science and Technology of China, Chengdu, Sichuan, China}
\mlsysaffiliation{upm}{Universidad Politécnica de Madrid, Madrid, Community of Madrid, Spain}

\mlsyscorrespondingauthor{Shanshan Liu}{ssliu@uestc.edu.cn}

\mlsyskeywords{Large Language Models, Reproducibility, LLM Serving, Batching, Prefill}

\vskip 0.3in

\begin{abstract}
Reproducible LLM inference requires that a request produces the same output regardless of the other requests in the batch. Existing solutions pay for this in different ways: batch-invariant kernels constrain every operator, and isolating prefill with fixed-shape decode ({our CoRun-style baseline}) gives up shared prefill compute and pads decode. We present BlockLock, which keeps shared batched prefill and pins only the numerical path of the protected request: it fixes the geometry that the request sees, with {two modes: BlockLock-Rows uses fixed-width rows and retains batched decode of all requests; BlockLock-Pack uses terminal-anchored packing with an isolated dummy segment, then extracts the protected request's KV state for target-only decode. Under a stated per-deployment contract, both modes preserve the target's output bitwise when its companions change.} Across five {dense} models (Qwen3-8B/14B, Llama-3.1-8B-Instruct, and Gemma-3-4B/12B-IT) and three single-GPU deployments (NVIDIA A6000, A100, and H100), we evaluate our approach on both tinyBenchmarks and BFCL{; we also include initial tensor-parallel experiments on two NVIDIA A6000 GPUs. Both modes preserve target sequences and Top20-logit bits in the reported tests.} At equal delivered work, {BlockLock-Rows} reduces latency by 53.5--55.0\% and improves throughput by 115.2--122.2\% relative to vLLM Batch-Invariant; compared with batch-invariant kernels, it needs no custom kernel implementations and constrains only the protected requests.
\end{abstract}
]

\printAffiliationsAndNotice{\mlsysEqualContribution} 

\section{Introduction}

Large language models (LLMs) are now used in evaluation, in reinforcement learning, and in agentic workflows where the model calls external tools \citep{maiapolo2024tinybenchmarks,deepseekai2025r1,patil2025bfcl}. In these uses, the same request should produce the same output. If the output changes between runs, the comparison of two models can change, the data used to train a policy is not the data that the policy would generate, and an agent can take a different action. Reproducibility is therefore important to trust the results of these systems.

In practice, current serving systems do not provide reproducibility. Even with greedy decoding and fixed seeds, the output of a request can change when the other requests in the batch change \citep{he2025defeating,fu2026beyond}. The reason is that floating point addition is not associative, so the result depends on the order of the sums. The order depends on the shape of the tensors that the kernels process, and the shape depends on the lengths of the other requests in the batch. A small change in the logits can flip a near-tie between two tokens, and from that point the rest of the generation is different \citep{fu2026beyond}. This effect appears in real workloads: in a study with 188 queries from the Berkeley Function Calling Leaderboard, two queries change the generated function call when the batch geometry changes (Section 2).

\begin{figure}[!t]
\centering
\includegraphics[width=\columnwidth]{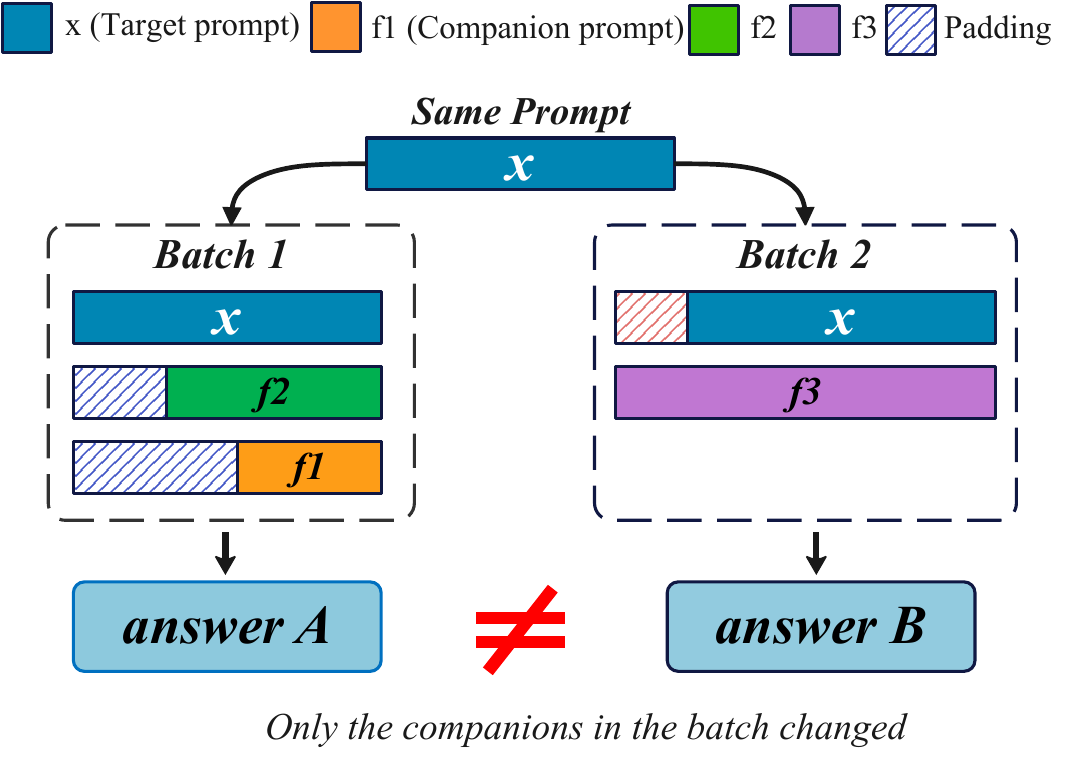}
\caption{Batch-induced nondeterminism. Processing the exact same target prompt $x$ (blue) alongside different companion requests ($f_1, f_2$  vs. $f_3$) alters tensor padding and reduction orders, flipping near-tie logits to produce divergent outputs.}
\label{fig:problem}
\end{figure}

At the same time, inference has to be efficient. Interactive services need low latency, and evaluation, rollout generation, and reasoning workloads need sustained throughput. Batching and scheduling are the main sources of efficiency in modern serving systems \citep{yu2022orca,kwon2023vllm,agrawal2024sarathi,zhong2024distserve}, together with optimized attention kernels \citep{dao2022flashattention}. A solution for reproducibility that removes batching is therefore expensive.

Existing solutions obtain reproducibility in different ways, and all of them have a cost. Batch-invariant kernels constrain the reduction order of every operator so that the result does not depend on the batch shape \citep{he2025defeating}. This requires specialized kernel implementations for each operator and each hardware, and the constraint is applied to all the requests, also to the ones that do not need reproducibility. Verified speculation runs a fast path that can be non-reproducible and then verifies the result and rolls back when it is not the canonical one \citep{gond2026llm42}; the verification and the rollback add work. CoRun isolates the prefill of each request, with batch size one, uses fixed-shape batched decode, and supports seeded sampling by binding the random-number-generator state to the request instead of the batch slot \citep{zhao2026corun}. The CoRun baseline is simple, but it gives up shared prefill compute, so the GPU processes one prompt at a time even when several prompts could be processed together. {Our CoRun-style baseline has 3.12--3.28$\times$ the end-to-end latency of shared batched execution (Section~4); this comparison does not isolate the cost of serialized prefill.} 

BlockLock takes a different approach. It keeps shared batched prefill, protects only the requests that need reproducibility, and runs on a stock framework without custom kernels or captured graphs. The main idea is that the numbers of a request stay fixed if the shape that the kernels see for that request stays fixed. BlockLock therefore pins the geometry of the protected request instead of removing the batch, with two schemes that fix the geometry in different ways: BlockLock-Rows and BlockLock-Pack. Companions share the same prefill invocation, but the protected request always sees the same tensor shapes, the same positions, and the same reduction order, independently of the contents and the lengths of the companions.

{This design is the opposite trade from batch-invariant kernels. Batch-invariant kernels guarantee determinism across different batch sizes, so every operator is constrained for every request. BlockLock instead fixes the batch size and the execution geometry for each deployment. When the geometry is fixed, every run executes the same kernels with the same shapes, the same positions, and the same reduction order, so the protected request produces bitwise identical outputs with the stock kernels. The batch size and the class widths are free parameters of the deployment: they are chosen to fit the hardware and the workload, and the determinism contract holds for that fixed configuration. 
{Fixing the geometry retains shared batched execution, subject to padding and backend costs:} the protected request shares one prefill invocation with its companions, and in BlockLock-Rows all the requests decode together in the normal batched decode. The cost of the fixed geometry is the padding of the rows or the filler segment, and Section 4 quantifies this cost.}

BlockLock combines deterministic length classes with two schemes. In the first scheme, BlockLock-Rows, the batch is a tensor of shape $[B, W]$ and the protected request keeps a stable row when the companions change. In the second scheme, BlockLock-Pack, the requests are packed into one block of shape $[1, BW]$, the protected request is anchored at the end of the block, and an isolated dummy segment absorbs the remaining space. After the prefill, the decode continues from the request's own KV state, so the numerical path of the protected request is preserved end to end. BlockLock does not need new kernels: it works with stock frameworks and only requires that the operators are repeat-stable for a fixed shape, which is checked as part of a per-deployment contract (Section 3 and Appendix A).

BlockLock-Rows retains shared $[B, W]$  prefill and decodes all $B$ requests simultaneously without padding decode to a worst-case server-wide concurrency envelope. Both modes still reserve $BW$ prefill positions.

This paper makes the following contributions:
\begin{enumerate}
\item \textbf{Batch-dependent output is a geometry problem, not only a kernel problem.} Padding, row width, and position change with the companions. Padded batching agrees only 52.59\% across companion changes, and 
{the default backend path can vary even at $B = 1$}.
\item \textbf{BlockLock fixes the target's execution geometry instead of the batch.} BlockLock-Rows cuts the prompts into blocks of size $L$ in fixed-width rows. BlockLock-Pack uses one block with the target at the end and a segment-causal mask. Both keep batched prefill and need no new kernels.
\item \textbf{Determinism after prefill under three conditions} (canonical geometry, repeat-stable arithmetic, faithful cache handoff) is validated with 100\% agreement over ten repeats and 100\% for single and double GPU implementations.
\item {\textbf{Measured efficiency on Torch and vLLM.} Across eight standard benchmarks, BlockLock-Rows incurs only 0.5--0.8\% latency overhead over standard PyTorch batching and runs 2.15--2.22$\times$ faster than vLLM Batch-Invariant (BI) at equal delivered work.}
\end{enumerate}

The rest of the paper is organized as follows. Section 2 gives the background and the related work. Section 3 describes the design, Section 4 presents the experimental results, and Section 5 discusses the scope and the limitations. The formal contract and the proof are in Appendix A, and the complete experimental details are in Appendix B.

\section{Background and Related Work}

\subsection{Prefill, Decode, and Cross-Phase KV Reuse}

Autoregressive inference has two distinct phases. Prefill processes prompt tokens in parallel and constructs a KV cache. Decode appends a generated token at each step and reuses that cache instead of recomputing the full history. Dynamic batching and efficient cache management improve utilization \citep{yu2022orca,kwon2023vllm}; chunked prefill additionally partitions a prompt across model invocations \citep{agrawal2024sarathi}.


{However, in standard dynamic batching, sharing a prefill invocation ties each request's tensor padding and reduction tree to the lengths of co-scheduled requests. The central systems challenge addressed by BlockLock is to preserve efficient inference while keeping the protected request's execution geometry invariant to changes in companion requests.}

\subsection{Repeat Variation and Configuration Variation}

Repeat variation occurs when equal logical inputs and execution geometry yield unequal numerical outputs in repeated executions. Execution-configuration variation occurs when changing $B$, padding, physical offsets, or a distributed configuration changes the arithmetic path, even if each individual configuration is repeat-stable. In Floating-point addition it can occur that $(a + b) + c \neq a + (b + c)$ because intermediate rounding depends on the reduction order. Shape-dependent kernel dispatch and tile decomposition can therefore change logits \citep{he2025defeating,fu2026beyond}. Batch invariance preserves request numerics across varying shapes. Canonical reproducibility assigns a target a repeatable plan within one deployment. Our backend comparisons characterize numerical behavior under controlled policies; identifying an internal mechanism such as atomic accumulation or split reduction requires kernel-level evidence.

\subsection{Backend Diagnosis and Autoregressive Amplification}

The problem is not only the batch: the backend itself can change the result. {In a baseline single-request audit at $B = 1$ (Appendix~C), the default backend dispatch produced multiple divergent sequences for the identical prompt, whereas enforcing our restricted backend policy yielded 100\% bitwise raw-logit agreement. Consequently, backend eligibility forms a foundational requirement of our reproducibility conditions (Section 3.4) }

The same effect appears under batching: across the {all configurations}, Padded batching reaches only 52.59\% pair-weighted generated-sequence agreement, while CoRun and both BlockLock schemes reach 100\% (Section 4 and Appendix B). Deterministic control is therefore needed already at $B = 1$, and batching makes the variation larger.

{On NVIDIA H100, we observed that the default cuDNN attention backend introduces numerical variations at $B = 1$, ultimately causing differences in output logits and generated token sequences. We evaluated Qwen3-8B using an identical MMLU prompt, a fixed batch size of $B=1$, and 320 decoding steps with greedy decoding. With cuDNN enabled, CoRun produced two distinct token sequences across four repeated runs produced three and two distinct sequences. After disabling cuDNN attention, all three schemes achieved bitwise-identical logits and token sequences across repeated executions. Further analysis revealed that small numerical perturbations could change greedy token selection: at decoding step 280, a logit value changed from 24.625 to 24.5, reversing the ranking of two competing tokens and causing subsequent output divergence.}

\subsection{Related Work}

\paragraph{Deterministic kernels.} Batch-invariant kernels constrain the reduction order of all the operators so the result does not depend on the batch shape \citep{he2025defeating}. This requires specialized kernel implementations for each operator and each hardware. BlockLock instead pins only the geometry of the protected request and works with stock kernels. {The two approaches answer different questions. 
{Batch-invariant kernels preserve request numerics across varying batch shapes. BlockLock permits companion contents and lengths to change within a fixed canonical plan, but does not guarantee equality across different batch sizes.} BlockLock fixes the batch size and the geometry of the compared runs, so it does not constrain any operator and uses the stock kernels of the framework.}

\paragraph{Retry on difference.} LLM-42 runs a fast path that can be non-reproducible, and then verifies the result and rolls back when it is not the canonical one \citep{gond2026llm42}. BlockLock does not need verification or rollback because the protected request follows the canonical path. Vosti specifies, implements, and verifies deterministic inference in a formal way \citep{qin2026vosti}; this is complementary to BlockLock, and the BlockLock contract could be a target for this kind of verification.

\begin{figure}[!t]
\centering
\includegraphics[width=\columnwidth]{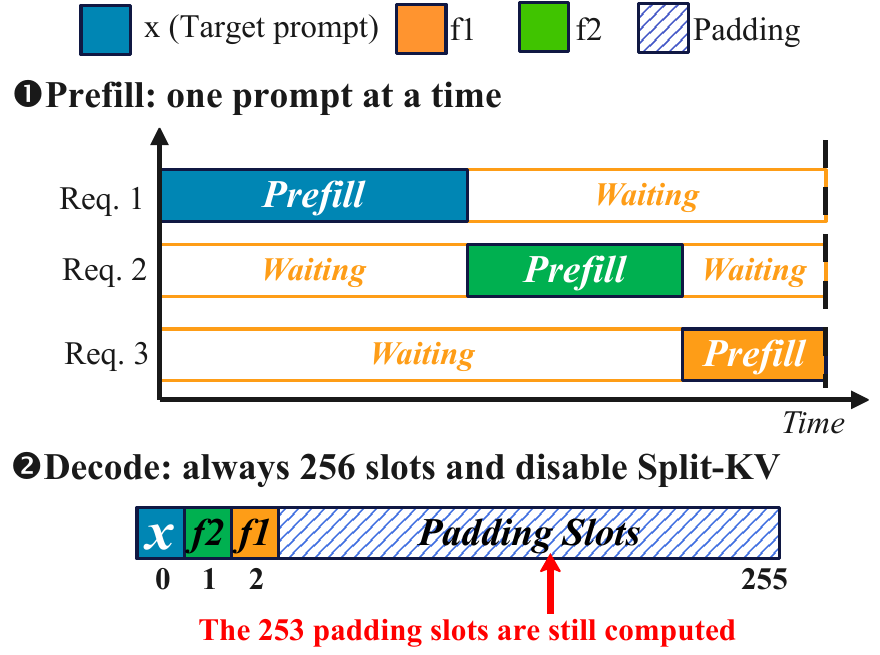}
\caption{Limitations of the CoRun-style baseline. (1) Prefill executes sequentially ($B=1$) to avoid cross-request interference, forfeiting batching efficiency. (2) Decode enforces a fixed 256-slot batch shape, wasting compute on masked padding slots when concurrency is low. {The 256-slot envelope depicts CoRun's evaluated configuration.}} 
\label{fig:corun}
\end{figure}

\paragraph{Isolated scheduling.} CoRun isolates the prefill of each request and uses fixed-shape batched decode \citep{zhao2026corun}, as Figure~\ref{fig:corun} illustrates. CoRun also covers seeded sampling with request-bound random-number state, and its decode path uses captured CUDA graphs. BlockLock keeps the companions in the same prefill invocation, pins only the target, and runs on a stock framework without CUDA graphs; the request-bound sampling technique of CoRun applies to BlockLock as well.

{CoRun's graph capacity is configurable: 256 matches its evaluated maximum concurrency \citep{zhao2026corun}. High occupancy amortizes padding; low occupancy wastes slots. Isolated prefill adds waiting behind earlier prompts. Our fixed-group timings do not measure queueing or tail latency.}

\paragraph{Padded and prepacked batching.} Prepacking packs prompts of different lengths into one batch to reduce padding and improve throughput \citep{zhao2025prepacking}; the packing of BlockLock instead anchors the protected request so that its geometry does not change. Fu et al. measure nondeterminism through token probabilities and relate it to the execution configuration \citep{fu2026beyond}; BlockLock acts on the cause, the shape that the kernels see, instead of measuring the effect.

\paragraph{Serving systems.} Serving systems improve efficiency with batching, cache management, chunked prefill, and prefill--decode disaggregation \citep{yu2022orca,kwon2023vllm,agrawal2024sarathi,zhong2024distserve}, and optimized attention kernels reduce the cost of each invocation \citep{dao2022flashattention}. {Length-based bucketing is already used in LLM serving systems to reduce padding overhead and improve resource utilization, as demonstrated by BucketServe\cite{bucketserve2025}.} These works do not consider reproducibility. In the measured setting, BlockLock keeps the shared batched prefill for the companions while it makes the protected request reproducible; Section 4 quantifies the cost of this protection.

\section{Design}

\begin{figure*}[!htbp]
\centering
\includegraphics[width=\textwidth]{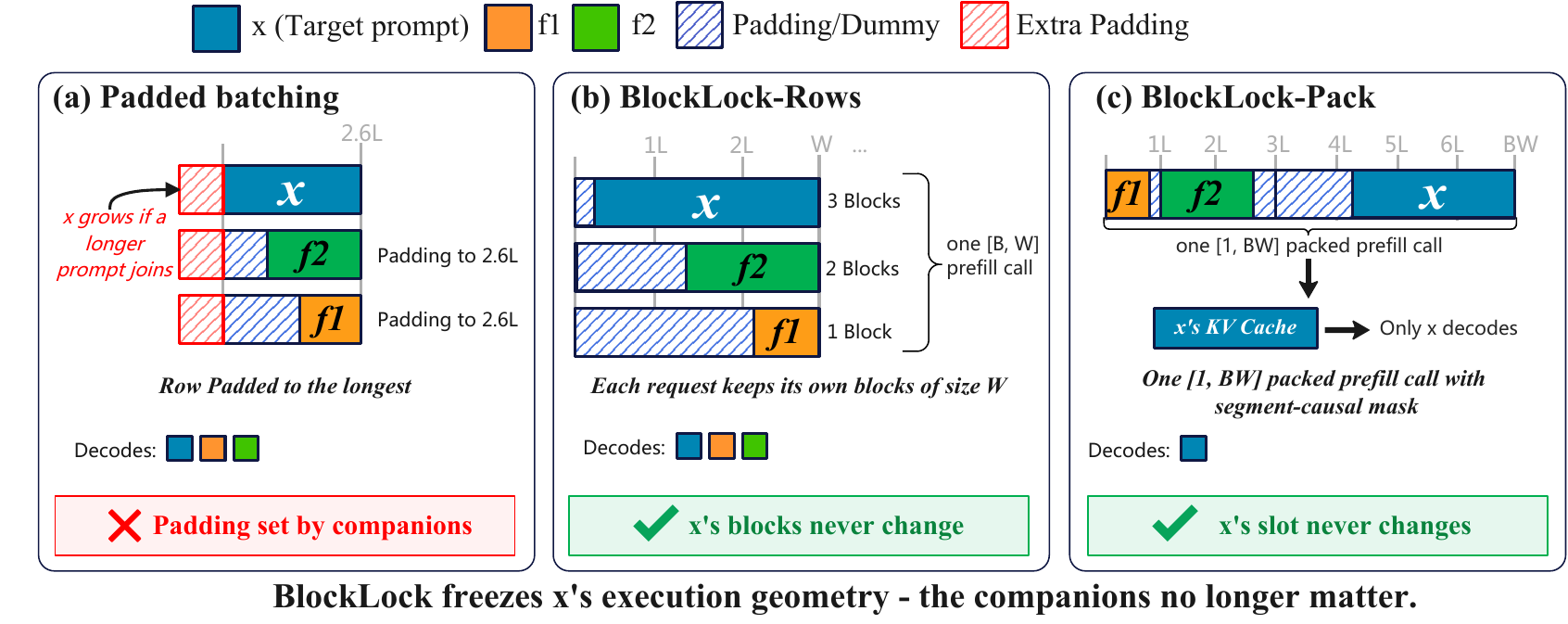}
\caption{Execution geometry comparison. (a) Padded batching pads all rows to the longest request in the batch, causing $x$’s padding offset to shift with its companions. (b) BlockLock-Rows executes a single $[B, W]$  prefill where every row is aligned to a fixed class width $W$ (a multiple of block size $L$), preserving $x$’s exact geometry and batched decode across all $B$ requests. (c) BlockLock-Pack concatenates requests into a $[1, BW]$  block with $x$ anchored at the terminal interval $[P-L_x, P)$  and an isolated dummy segment absorbing length variations, followed by target-only KV extraction and decode.}
\label{fig:schemes}
\end{figure*}

BlockLock coordinates the execution geometry, the token layout, the backend eligibility, and the distributed execution so that the numerical path of a protected request is preserved through prefill and decode. This section gives the main ideas. Figure~\ref{fig:schemes} shows the two schemes on one example. {The batch size and the geometry are fixed only for the runs that are compared; each deployment chooses them freely for its hardware and its workload.}

\subsection{Setup: one protected request and many companions}

The workload has one protected target request $x$, whose output must be reproducible, and ordinary companion requests that share the GPU work. {BlockLock's scheduler assigns tokenized lengths to fixed classes using an alignment granularity $L$ {(set to $L = 64$ tokens in our standard evaluation}) and merges adjacent buckets when needed. Then prompts} are cut into blocks of size $L$, and a prompt of arbitrary length occupies the blocks it needs, so the width $W$ of its class is a multiple of $L$. The class mapping is frozen for the executions that are compared. A group of $B$ requests of one class reserves $P = B W$ prompt positions. The group contains the target and $B-1$ companions, and a seeded schedule changes the contents and the lengths of the companions between repetitions while the target, $B$, and $W$ stay fixed.

We compare against two baselines: Padded batching, the normal serving behavior without any reproducibility control; CoRun, which isolates the prefill of each request and uses fixed-shape batched decode \citep{zhao2026corun}.

\subsection{BlockLock-Rows: fixed-width rows}

BlockLock-Rows processes each group as a single tensor of shape $[B, W]$. {To keep serving efficiency and reduce padding overhead, it first groups incoming prompts into length-based buckets.} Within each bucket, each request occupies a fixed-width row, with left padding aligning token sequences to the bucket width and an attention mask distinguishing real tokens from padding. The target keeps the same row and the same logical positions when the companions change, so the kernels always see the same shape for the target. One prefill invocation processes all the rows, and the decode continues with the normal batched KV cache of the $B$ requests. {Because each row in the $[B, W]$ tensor occupies a fixed width $W$ and undergoes independent row-wise reductions, the reproducibility contract (Section~3.4) holds symmetrically for any batch slot assigned a canonical plan. In our evaluation protocol (Section~4), we isolate one slot as the fixed target $x$ while intentionally varying the contents and lengths of the other $B - 1$ companion rows across repetitions to rigorously stress-test cross-request invariance. For dense full attention, the score computation takes $O(BW^2)$ work. BlockLock-Rows does not need an explicit mask between segments, which is its main cost advantage over BlockLock-Pack.}

\subsection{BlockLock-Pack: terminal-anchored packing}

BlockLock-Pack concatenates the companions, an optional dummy segment, and the target along the sequence dimension into one block of shape $[1, P]$. The capacity that is not used by real tokens is filled by an isolated dummy segment, which absorbs the changes in the companion lengths. The target is always placed last, so its physical interval is $I_x = [P - L_x, P)$ and both endpoints stay fixed when the companions change. For example, with $W = 192$ and $B = 2$, the block reserves 384 positions, and companion and target lengths of 180 and 192 leave 12 dummy positions without moving the target.

Each segment uses local positions $0, 1, \ldots, L_i - 1$, and a segment-causal mask allows attention only inside the same segment. One forward call processes the whole packed block and produces the packed KV cache and the logits of the last position of the target. After the prefill, BlockLock-Pack extracts the KV state of the target layer by layer: full-attention layers keep the last $L_x$ key and value entries, and sliding-attention layers keep the visible target suffix and restore its logical cumulative length. The decode then continues from this target-local cache, one token per step, so the target attends only to its own state. 
{In native PyTorch, the packed path uses a dense segment mask with $O(P^2)$ storage and score work; the vLLM integration uses paged-attention metadata instead (Appendix A.2);} in exchange, the target decodes alone and does not pay for a batch it does not need.

BlockLock-Pack protects the prompt anchored at the end of the block, and it is the option for low-concurrency serving: protecting several prompts needs one pack per protected prompt, which costs extra prefill. {By design, BlockLock-Pack targets low-concurrency or single-audit scenarios where only the terminal request $x$ requires decoding. Attempting to batch-decode the companions alongside $x$ after unpadded KV extraction would expose the decode attention kernel to the companions' varying sequence lengths $L_i$, violating canonical geometry during decode (Whereas padding all KV caches back to $W$ simply reduces to BlockLock-Rows). Isolating $x$’s KV state for single-request decoding guarantees that both prefill and decode geometries remain strictly invariant to companion lengths.}

\subsection{What makes the output reproducible}\label{sec:repro}

The two schemes give the same contract. For a fixed deployment (hardware, software versions, precision, backend eligibility, length-class mapping, $B$, and the chosen scheme), the raw logits of the target are bitwise identical at every prediction index, for any companions. Three conditions are sufficient:
{\renewcommand{\labelenumi}{C\arabic{enumi}.}
\begin{enumerate}
\item Canonical geometry: the geometry and the logical isolation of the target are preserved.
\item Repeat-stable arithmetic: the operators and the distributed collectives are repeat-stable.
\item Faithful handoff: the cache handoff is faithful.
\end{enumerate}} The backend policy excludes the components that are not repeat-stable (in our experiments, cuDNN SDPA and TF32), and each deployment is validated before the comparison. The contract is stated for greedy decoding; seeded sampling fits the same design with request-local random-number state, the technique that CoRun uses for its request-bound sampling \citep{zhao2026corun}. The formal statement, the sufficient conditions, and the proof are in Appendix A.

The same logical plan runs on one GPU or on one model tensor-parallelized across two A6000 GPUs. In the two-GPU case, one rank constructs and broadcasts the plan, the ranks check that their input plans agree before inference, and the collectives must also be repeat-stable.

\subsection{Overhead and scaling}

The overhead of the two schemes comes from three sources. The first is the reserved space: both schemes reserve $P = B W$ positions, so BlockLock-Rows pads each row to the class width $W$, and BlockLock-Pack fills the unused capacity with the dummy segment, which is wasted work that grows when the companions are shorter than the reserved space. The second is the batch size: the packed mask costs $O(P^2) = O(B^2 W^2)$ storage and score work, so it grows quadratically with $B$, while the rows keep $O(BW^2)$ attention work, linear in $B$. The third is the prompt length: longer prompts move to a wider class, which increases the reserved width and the padding of the whole class. 
{Across the eight-dataset equal-delivered-work benchmark at $B=4$ and 64 output tokens per request, the geometric-mean latency overhead of Torch Rows over Torch Padded is 0.5--0.8\%}. For distributed execution ($TP = 2$), our evaluation focuses on verifying numerical invariance across multi-GPU collective communications, confirming 100\% bitwise agreement.


\section{Experimental Results}

This section presents the evaluation results in 
{terms} of determinism and performance. The exact model names, links, software versions, and the full protocol are in Appendix B.

\subsection{Experimental setup}

The evaluation covers three GPU platforms on two architectures: A6000 and A100 (Ampere) and H100 (Hopper). The A6000 configurations include single-GPU experiments and a separate experiment that partitions one model across two A6000 GPUs with tensor parallelism (TP=2). The main models are Qwen3-8B and Qwen3-14B, Llama3.1-8B-Instruct, Gemma3-4B and Gemma3-12B-it. The inputs are seven tinyBenchmarks \citep{maiapolo2024tinybenchmarks} and BFCL multiple-function queries \citep{patil2025bfcl}. All runs use BF16 precision and greedy decoding across batch sizes $B \in \{1, 2, 4, 8, 16\}$. Each target and batch configuration is evaluated across 10 independent repetitions with distinct, randomly sampled companion combinations, yielding $\binom{10}{2} = 45$ pairwise cross-companion comparisons per condition (60,120 pairwise comparisons across all 1,336 evaluated conditions). Exact model weights, repository links, and software versions are listed in Appendix~B.

\subsection{Checking determinism}

\begin{figure*}[!htbp]
\centering
\includegraphics[width=\textwidth]{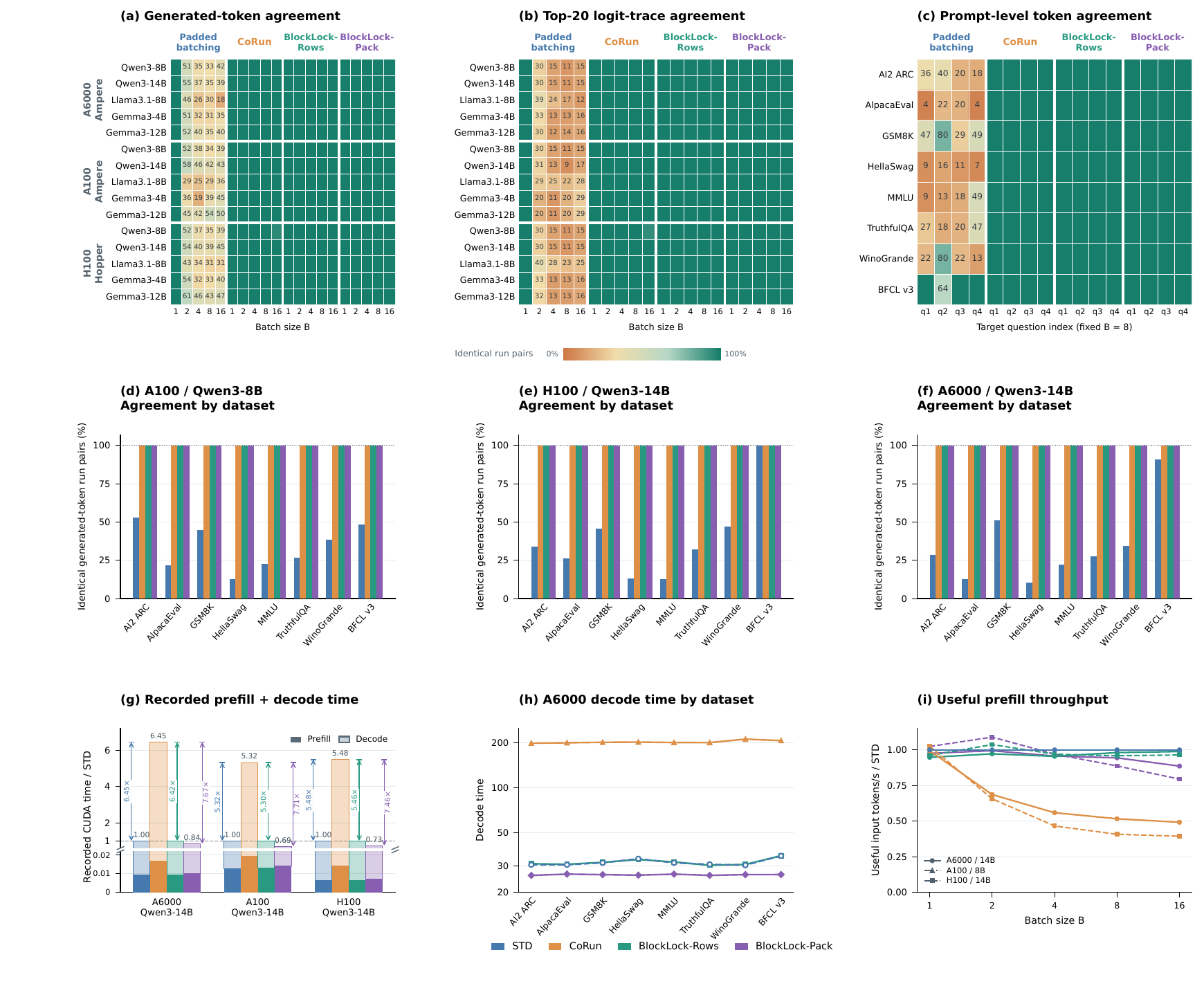}
    \caption{
{Reproducibility and phase efficiency across GPUs and models. (a,b) Generated-token and Top-20 logit-trace agreement within each fixed batch size. (c) Per-target token agreement. (d--f) Token agreement by dataset. (g) Prefill and decode time normalized to STD. (h) Per-dataset decode latency (evaluated at 500 output tokens per request). (i) Useful-input prefill throughput relative to STD.}
    }
    \label{fig:performance}
\end{figure*}

{The first check is the backend itself. We first verify at $B = 1$ on each hardware that all tested runs reach 100\% agreement on both Top-20 logits and output tokens. For each target, we repeat the same request with different companions and compare the outputs at both the logit and output-token levels.} Figure~\ref{fig:performance}(a,b) summarizes generated-token and Top-20 logit-trace agreement across three hardware platforms, five models, and batch sizes $B\in\{1,2,4,8,16\}$. Agreement is the fraction of identical run pairs within each target--configuration condition: $R$ repetitions provide $\binom{R}{2}$ pairs, including 45 pairs for ten repetitions. Token agreement compares the actual generated token-ID sequences; Top-20 agreement requires identical candidate token IDs and raw-logit values at every generation step. Figure~\ref{fig:performance}(a,b) pool pairs within each dataset and then average dataset-level agreement equally. Figure~\ref{fig:performance}(c) exposes variation across individual targets on A6000/Qwen3-14B at $B=8$, showing q1--q4 separately for each dataset and policy. Figure~\ref{fig:performance}(d--f) report agreement by dataset at $B=8$ for A100/Qwen3-8B, H100/Qwen3-14B, and A6000/Qwen3-14B, respectively. Each dataset contributes four targets with ten repetitions each, yielding 180 comparison pairs per policy.

The main result is that both schemes reproduce the target exactly. {Both BlockLock-Rows and BlockLock-Pack reach 100\% target-sequence and Top20 agreement across single-GPU conditions and TP conditions. 
{Appendix B.6 reports 52.59\% sequence and 35.93\% Top20 agreement for the single-GPU subset; the TP conditions are reported separately.} The vLLM BI and all four BlockLock implementations reach full agreement in every one of their 80 conditions. Standard batching has dataset-dependent variation; some conditions also agree fully. These comparisons hold $B$ and the deployment fixed.}

The two-GPU experiment shows the same behavior. With one Gemma3-4B model partitioned across two A6000 GPUs (TP=2), both schemes reach 100\% agreement in their 20 tested conditions each, while Padded batching degrades from 100\% at $B = 1$ to 72.4\%, 43.6\%, and 21.8\% at $B = 2, 4, 8$.

\subsection{Performance: latency and throughput}

\begin{table*}[!htbp]
\centering
\caption{Phase latency and throughput comparison across eight datasets on NVIDIA A6000. 
{At $B=4$, all schemes decode four requests.}}
\label{tab:target-protocol}
\footnotesize
\setlength{\tabcolsep}{6pt}
\begin{tabular}{llrrrrrr}
\toprule
Model & Method & Prefill (ms) & Decode (ms) & Total (ms) & Speedup  & Output tokens & Tokens/s \\
\midrule
Qwen3-8B & Padded batching & 66.99 & 1213.59 & 1280.58 $\pm$ 4.17 & 3.362$\times$ & 256 & 199.91 \\
 & CoRun  & 91.55 & 4214.20 & 4305.75 $\pm$ 18.27 & 1.000$\times$ & 256 & 59.46 \\
 & BlockLock-Rows & 76.58 & 1236.13 & 1312.71 $\pm$ 3.83 & 3.280$\times$ & 256 & 195.02 \\
 & BlockLock-Pack & 83.86 & 1110.04 & 1193.91 $\pm$ 10.37 & 3.606$\times$ & 64 & 53.61 \\
\midrule
Qwen3-14B & Padded batching & 106.16 & 1756.99 & 1863.16 $\pm$ 2.32 & 3.211$\times$ & 256 & 137.40 \\
 & CoRun  & 137.02 & 5846.13 & 5983.14 $\pm$ 0.43 & 1.000$\times$ & 256 & 42.79 \\
 & BlockLock-Rows & 123.53 & 1790.02 & 1913.56 $\pm$ 1.59 & 3.127$\times$ & 256 & 133.78 \\
 & BlockLock-Pack & 133.94 & 1517.66 & 1651.60 $\pm$ 1.73 & 3.623$\times$ & 64 & 38.75 \\
\bottomrule
\end{tabular}
\end{table*}

{
\textbf{Phase Cost Characterization.} We instrument prefill and decode intervals to evaluate the efficiency of both BlockLock primitives across PyTorch and vLLM engines. Because BlockLock-Rows preserves the standard $[B, W]$ batched KV cache to decode all $B$ requests simultaneously, whereas BlockLock-Pack extracts only the terminal target's KV state to decode a single request after a shared $[1, BW]$ prefill, we explicitly distinguish between equal-delivered-work comparisons ($B$ requests decoded) and single-target audit workloads.}

{
\textbf{Comparison Against Sequential Isolation (CoRun-style).} Figure~\ref{fig:performance}(d--f) and Table~\ref{tab:target-protocol} compare BlockLock against standard padded batching (STD) and sequential request isolation (CoRun). As shown in Table~\ref{tab:target-protocol}, {the measured CoRun-style baseline takes $4305.75$~ms on Qwen3-8B and $5983.14$~ms on Qwen3-14B ($B=4$).} 
In contrast, BlockLock-Rows completes in $1312.71$~ms and $1913.56$~ms, achieving a 3.28$\times$ and 3.13$\times$ end-to-end speedup (and $195.02$ vs.\ $59.46$~tokens/s on Qwen3-8B) 
{with 2.5\%--2.7\% higher total latency than Padded batching in these two rows of Table~\ref{tab:target-protocol}.} Across all seven tinyBenchmarks on A6000, A100, and H100 (Figure~\ref{fig:performance}(f)), BlockLock-Rows sustains nearly $100\%$ of standard useful-input prefill throughput up to $B=16$.}

Decoupling the phases in Table~\ref{tab:target-protocol} clarifies two distinct sources of efficiency gain: during prefill, sharing the $[B, W]$  invocation avoids $B = 1$ serialization; during decode, scoping the canonical batch geometry to the active $B = 4$ cohort avoids computing across CoRun's fixed slot server-wide decode envelope.

\begin{figure*}[!htbp]
\centering
\includegraphics[width=\textwidth]{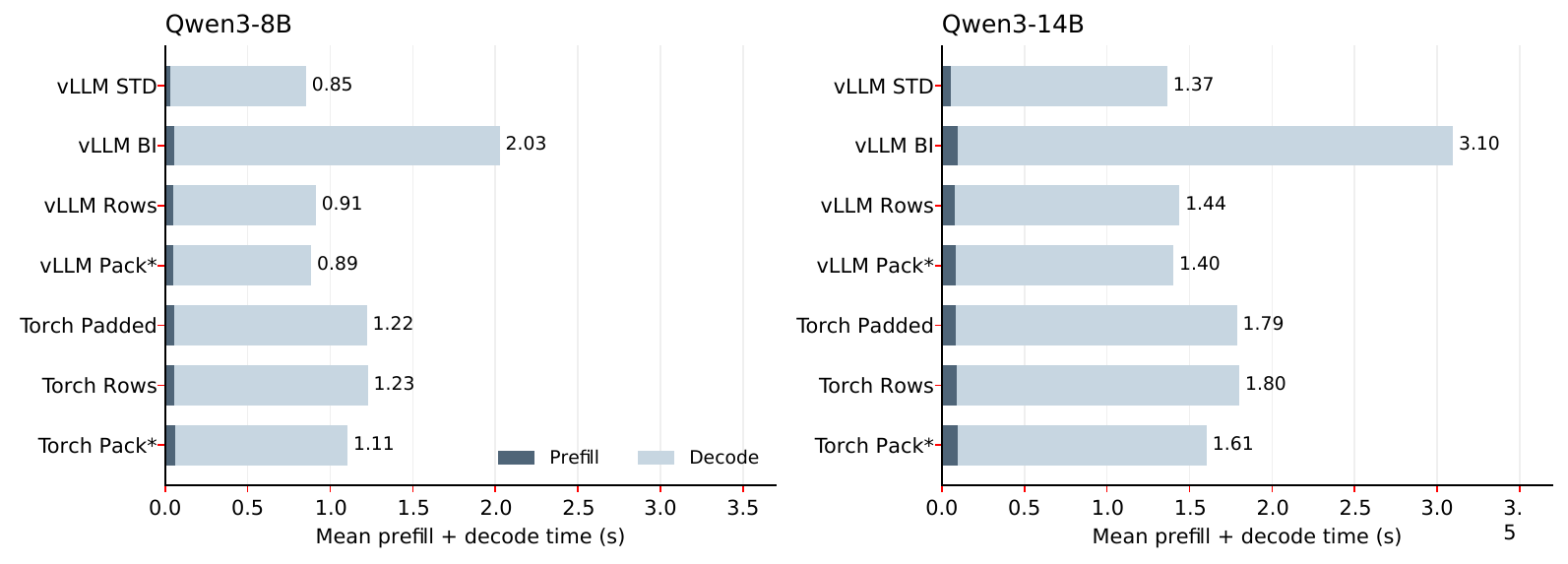}
\caption{{Mean instrumented prefill and decode latency across the seven evaluated methods, averaged over eight datasets ($B=4$, 64 output tokens per decoded request). Dark and light bars show prefill and decode. Asterisks mark target-only Pack (one output request); other methods decode four. }}
\label{fig:perf}
\end{figure*}

{
\textbf{Comparison Against Batch-Invariant Kernels (vLLM BI).} Figure~\ref{fig:perf} and Table~\ref{tab:seven-stacked} 
{evaluate seven implementations across eight datasets in two workload scopes ($B=4$): STD, BI, Padded, and Rows deliver 256 output tokens per group, while Pack delivers only the protected target's 64 output tokens. Equal-delivered-work speedups compare Rows with full-batch baselines; Pack timings describe a different, target-only workload.} Because batch-invariant kernels (vLLM BI) constrain reduction orders across all operators, they incur more than a $2\times$ latency penalty over standard vLLM (vLLM STD). By pinning the execution geometry rather than constraining kernel arithmetic, vLLM Rows incurs only a $5.4\%$--$6.9\%$ geometric-mean latency overhead over vLLM STD across the eight datasets, reducing total execution time by 55.0\% (Qwen3-8B) and 53.5\% (Qwen3-14B) and boosting output throughput by 122.2\% and 115.2\% relative to vLLM BI. In native PyTorch, Torch Rows adds merely $0.5\%$--$0.8\%$ overhead over Torch Padded. Meanwhile, when only the audited target's output is needed, Pack further reduces total wall-clock time by avoiding companion decode overhead ($0.89$~s vs.\ $0.91$~s in vLLM; $1.11$~s vs.\ $1.23$~s in PyTorch on Qwen3-8B).}


\begin{table*}[!htbp]
\centering
{
\caption{Means of Prefill+decode time (ms). 
{$B=4$; full-batch methods generate 256 output tokens per group.}
{Relative latency changes below the times use vLLM STD or Torch Padded as the within-engine baseline. Positive values mean higher latency: $100(t/t_{\mathrm{base}}-1)\%$. N/A denotes a decrease.} 
}
\label{tab:seven-stacked}
\footnotesize
\setlength{\tabcolsep}{4pt}
\begin{tabular}{lrrrrrrr}
\toprule
& \multicolumn{4}{c}{vLLM} & \multicolumn{3}{c}{Torch} \\
\cmidrule(lr){2-5}\cmidrule(lr){6-8}
Dataset & STD & BI & Rows & Pack & Padded & Rows & Pack \\
\midrule
\multicolumn{8}{c}{\textbf{Qwen3-8B}} \\
MMLU & 871.32 & \begin{tabular}[t]{@{}r@{}}2032.84\\{(+133.3\%)}\end{tabular} & \begin{tabular}[t]{@{}r@{}}902.68\\{(+3.6\%)}\end{tabular} & \begin{tabular}[t]{@{}r@{}}874.84\\{(N/A)}\end{tabular} & 1194.34 & \begin{tabular}[t]{@{}r@{}}1201.19\\{(+0.6\%)}\end{tabular} & \begin{tabular}[t]{@{}r@{}}1101.97\\{(N/A)}\end{tabular} \\ 
AI2 ARC & 820.51 & \begin{tabular}[t]{@{}r@{}}2006.00\\{(+144.5\%)}\end{tabular} & \begin{tabular}[t]{@{}r@{}}884.39\\{(+7.8\%)}\end{tabular} & \begin{tabular}[t]{@{}r@{}}857.39\\{(N/A)}\end{tabular} & 1134.29 & \begin{tabular}[t]{@{}r@{}}1133.47\\{(-0.1\%)}\end{tabular} & \begin{tabular}[t]{@{}r@{}}1068.36\\{(N/A)}\end{tabular} \\ 
HellaSwag & 901.02 & \begin{tabular}[t]{@{}r@{}}2049.74\\{(+127.5\%)}\end{tabular} & \begin{tabular}[t]{@{}r@{}}952.99\\{(+5.8\%)}\end{tabular} & \begin{tabular}[t]{@{}r@{}}925.91\\{(N/A)}\end{tabular} & 1342.03 & \begin{tabular}[t]{@{}r@{}}1353.24\\{(+0.8\%)}\end{tabular} & \begin{tabular}[t]{@{}r@{}}1158.08\\{(N/A)}\end{tabular} \\ 
GSM8K & 845.56 & \begin{tabular}[t]{@{}r@{}}2000.99\\{(+136.6\%)}\end{tabular} & \begin{tabular}[t]{@{}r@{}}901.43\\{(+6.6\%)}\end{tabular} & \begin{tabular}[t]{@{}r@{}}872.76\\{(N/A)}\end{tabular} & 1179.14 & \begin{tabular}[t]{@{}r@{}}1194.50\\{(+1.3\%)}\end{tabular} & \begin{tabular}[t]{@{}r@{}}1084.28\\{(N/A)}\end{tabular} \\ 
TruthfulQA & 835.93 & \begin{tabular}[t]{@{}r@{}}2015.86\\{(+141.2\%)}\end{tabular} & \begin{tabular}[t]{@{}r@{}}883.42\\{(+5.7\%)}\end{tabular} & \begin{tabular}[t]{@{}r@{}}856.38\\{(N/A)}\end{tabular} & 1134.94 & \begin{tabular}[t]{@{}r@{}}1133.10\\{(-0.2\%)}\end{tabular} & \begin{tabular}[t]{@{}r@{}}1066.34\\{(N/A)}\end{tabular} \\ 
WinoGrande & 822.79 & \begin{tabular}[t]{@{}r@{}}1998.76\\{(+142.9\%)}\end{tabular} & \begin{tabular}[t]{@{}r@{}}883.49\\{(+7.4\%)}\end{tabular} & \begin{tabular}[t]{@{}r@{}}856.27\\{(N/A)}\end{tabular} & 1134.20 & \begin{tabular}[t]{@{}r@{}}1133.58\\{(-0.1\%)}\end{tabular} & \begin{tabular}[t]{@{}r@{}}1069.19\\{(N/A)}\end{tabular} \\ 
AlpacaEval & 803.92 & \begin{tabular}[t]{@{}r@{}}2005.83\\{(+149.5\%)}\end{tabular} & \begin{tabular}[t]{@{}r@{}}883.77\\{(+9.9\%)}\end{tabular} & \begin{tabular}[t]{@{}r@{}}856.41\\{(N/A)}\end{tabular} & 1132.41 & \begin{tabular}[t]{@{}r@{}}1136.67\\{(+0.4\%)}\end{tabular} & \begin{tabular}[t]{@{}r@{}}1063.79\\{(N/A)}\end{tabular} \\ 
BFCL v3 & 933.56 & \begin{tabular}[t]{@{}r@{}}2108.74\\{(+125.9\%)}\end{tabular} & \begin{tabular}[t]{@{}r@{}}1012.31\\{(+8.4\%)}\end{tabular} & \begin{tabular}[t]{@{}r@{}}985.65\\{(N/A)}\end{tabular} & 1530.01 & \begin{tabular}[t]{@{}r@{}}1542.91\\{(+0.8\%)}\end{tabular} & \begin{tabular}[t]{@{}r@{}}1233.03\\{(N/A)}\end{tabular} \\ 
\midrule
\multicolumn{8}{c}{\textbf{Qwen3-14B}} \\
MMLU & 1350.33 & \begin{tabular}[t]{@{}r@{}}3093.99\\{(+129.1\%)}\end{tabular} & \begin{tabular}[t]{@{}r@{}}1420.40\\{(+5.2\%)}\end{tabular} & \begin{tabular}[t]{@{}r@{}}1384.39\\{(N/A)}\end{tabular} & 1743.51 & \begin{tabular}[t]{@{}r@{}}1759.06\\{(+0.9\%)}\end{tabular} & \begin{tabular}[t]{@{}r@{}}1582.32\\{(N/A)}\end{tabular} \\ 
AI2 ARC & 1337.29 & \begin{tabular}[t]{@{}r@{}}3056.86\\{(+128.6\%)}\end{tabular} & \begin{tabular}[t]{@{}r@{}}1391.29\\{(+4.0\%)}\end{tabular} & \begin{tabular}[t]{@{}r@{}}1354.72\\{(N/A)}\end{tabular} & 1655.48 & \begin{tabular}[t]{@{}r@{}}1668.12\\{(+0.8\%)}\end{tabular} & \begin{tabular}[t]{@{}r@{}}1544.64\\{(N/A)}\end{tabular} \\ 
HellaSwag & 1426.65 & \begin{tabular}[t]{@{}r@{}}3140.35\\{(+120.1\%)}\end{tabular} & \begin{tabular}[t]{@{}r@{}}1508.24\\{(+5.7\%)}\end{tabular} & \begin{tabular}[t]{@{}r@{}}1471.27\\{(N/A)}\end{tabular} & 1968.27 & \begin{tabular}[t]{@{}r@{}}1988.79\\{(+1.0\%)}\end{tabular} & \begin{tabular}[t]{@{}r@{}}1690.10\\{(N/A)}\end{tabular} \\ 
GSM8K & 1344.17 & \begin{tabular}[t]{@{}r@{}}3048.39\\{(+126.8\%)}\end{tabular} & \begin{tabular}[t]{@{}r@{}}1420.78\\{(+5.7\%)}\end{tabular} & \begin{tabular}[t]{@{}r@{}}1383.55\\{(N/A)}\end{tabular} & 1736.45 & \begin{tabular}[t]{@{}r@{}}1762.16\\{(+1.5\%)}\end{tabular} & \begin{tabular}[t]{@{}r@{}}1578.48\\{(N/A)}\end{tabular} \\ 
TruthfulQA & 1328.39 & \begin{tabular}[t]{@{}r@{}}3058.07\\{(+130.2\%)}\end{tabular} & \begin{tabular}[t]{@{}r@{}}1391.19\\{(+4.7\%)}\end{tabular} & \begin{tabular}[t]{@{}r@{}}1354.85\\{(N/A)}\end{tabular} & 1660.74 & \begin{tabular}[t]{@{}r@{}}1667.44\\{(+0.4\%)}\end{tabular} & \begin{tabular}[t]{@{}r@{}}1541.07\\{(N/A)}\end{tabular} \\ 
WinoGrande & 1322.82 & \begin{tabular}[t]{@{}r@{}}3056.79\\{(+131.1\%)}\end{tabular} & \begin{tabular}[t]{@{}r@{}}1391.29\\{(+5.2\%)}\end{tabular} & \begin{tabular}[t]{@{}r@{}}1354.75\\{(N/A)}\end{tabular} & 1660.18 & \begin{tabular}[t]{@{}r@{}}1667.43\\{(+0.4\%)}\end{tabular} & \begin{tabular}[t]{@{}r@{}}1543.16\\{(N/A)}\end{tabular} \\ 
AlpacaEval & 1322.88 & \begin{tabular}[t]{@{}r@{}}3056.16\\{(+131.0\%)}\end{tabular} & \begin{tabular}[t]{@{}r@{}}1391.22\\{(+5.2\%)}\end{tabular} & \begin{tabular}[t]{@{}r@{}}1354.69\\{(N/A)}\end{tabular} & 1660.29 & \begin{tabular}[t]{@{}r@{}}1667.11\\{(+0.4\%)}\end{tabular} & \begin{tabular}[t]{@{}r@{}}1542.75\\{(N/A)}\end{tabular} \\ 
BFCL v3 & 1497.90 & \begin{tabular}[t]{@{}r@{}}3258.17\\{(+117.5\%)}\end{tabular} & \begin{tabular}[t]{@{}r@{}}1605.74\\{(+7.2\%)}\end{tabular} & \begin{tabular}[t]{@{}r@{}}1568.30\\{(N/A)}\end{tabular} & 2222.03 & \begin{tabular}[t]{@{}r@{}}2242.34\\{(+0.9\%)}\end{tabular} & \begin{tabular}[t]{@{}r@{}}1818.02\\{(N/A)}\end{tabular} \\ 

\bottomrule
\end{tabular}
}
\end{table*}

\begin{table*}[!htbp]
\centering{
\caption{Fixed-target workload time (s). 
{$B=4$. Padded, CoRun, and Rows decode all four requests; Pack decodes the protected target only. Each trace workload is one representative four-request group.}
{Relative latency changes below the times use Padded as the baseline. Positive values mean higher latency: $100(t/t_{\mathrm{base}}-1)\%$. N/A denotes a decrease.} 
}
\label{tab:workload-results}
\footnotesize
\setlength{\tabcolsep}{5pt}
\begin{tabular}{lrrrr}
\toprule
Width $W$ / output $T$ & Padded & CoRun & Rows & Pack \\
\midrule
1024 / 128 & 3.542 & \begin{tabular}[t]{@{}r@{}}8.472\\{(+139.2\%)}\end{tabular} & \begin{tabular}[t]{@{}r@{}}3.761\\{(+6.2\%)}\end{tabular} & \begin{tabular}[t]{@{}r@{}}2.654\\{(N/A)}\end{tabular} \\ 
1024 / 2048 & 72.988 & \begin{tabular}[t]{@{}r@{}}134.345\\{(+84.1\%)}\end{tabular} & \begin{tabular}[t]{@{}r@{}}75.812\\{(+3.9\%)}\end{tabular} & \begin{tabular}[t]{@{}r@{}}36.678\\{(N/A)}\end{tabular} \\ 
4096 / 128 & 8.187 & \begin{tabular}[t]{@{}r@{}}9.016\\{(+10.1\%)}\end{tabular} & \begin{tabular}[t]{@{}r@{}}9.176\\{(+12.1\%)}\end{tabular} & \begin{tabular}[t]{@{}r@{}}6.132\\{(N/A)}\end{tabular} \\ 
4096 / 2048 & 129.529 & \begin{tabular}[t]{@{}r@{}}135.103\\{(+4.3\%)}\end{tabular} & \begin{tabular}[t]{@{}r@{}}141.141\\{(+9.0\%)}\end{tabular} & \begin{tabular}[t]{@{}r@{}}39.669\\{(N/A)}\end{tabular} \\ 
Blazedit / 64 & 2.133 & \begin{tabular}[t]{@{}r@{}}4.421\\{(+107.3\%)}\end{tabular} & \begin{tabular}[t]{@{}r@{}}3.061\\{(+43.5\%)}\end{tabular} & \begin{tabular}[t]{@{}r@{}}2.339\\{(N/A)}\end{tabular} \\ 
BurstGPT / 64 & 1.173 & \begin{tabular}[t]{@{}r@{}}4.175\\{(+255.9\%)}\end{tabular} & \begin{tabular}[t]{@{}r@{}}1.198\\{(+2.1\%)}\end{tabular} & \begin{tabular}[t]{@{}r@{}}1.134\\{(N/A)}\end{tabular} \\ 
\bottomrule
\end{tabular}}
\end{table*}

\begin{table*}[!t]
\centering
\caption{Qualitative comparison of reproducible LLM inference approaches. BlockLock guarantees bitwise-identical outputs per deployment with repeat-stable kernels without requiring custom operator implementations.}
\label{tab:comparison}
\footnotesize
\setlength{\tabcolsep}{5pt}
\begin{tabular}{@{}p{0.11\textwidth}p{0.19\textwidth}p{0.12\textwidth}p{0.15\textwidth}p{0.13\textwidth}p{0.15\textwidth}@{}}
\toprule
Approach & Bitwise-reproducible  across companions & Needs custom  kernels & Batched prefill  efficiency & Decode  efficiency & Scope / limits \\
\midrule
Padded batching & No (52.59\% overall) & No & Yes (pads to longest) & Shared batched & No determinism guarantee \\
Batch-invariant kernels & Yes (across batch shapes) & Yes (per op/GPU) & Shared, constrained & Shared, constrained & Constrains all requests \\
LLM-42 & Yes (via verification) & No & Fast path + verify & Rollback on mismatch & Verification/rollback cost \\
CoRun & Yes (never mixed) & No & No ($B=1$ prefill) & Fixed-shape padded slots & Underutilizes prefill \\
\textbf{BlockLock-Rows} & \textbf{Yes (per deployment)} & \textbf{No} & \textbf{Shared $[B, W]$} & \textbf{Batched ($B$ requests)} & \textbf{Per deployment} \\
\textbf{BlockLock-Pack} & \textbf{Yes (per deployment)} & \textbf{No} & \textbf{Shared $[1, BW]$} & \textbf{Target only ($1$ request)} & \textbf{Per deployment} \\
\bottomrule
\end{tabular}
\end{table*}

{\textbf{Longer capacities and representative workloads.} Table~3 extends our controlled target-audit protocol to long-context capacities and extended generation lengths in the native PyTorch backend with full-logit verification. We report the mean of three interleaved CUDA timing trials per method and condition at $B = 4$. Four synthetic stress-test conditions evaluate class widths $W \in \{1024, 4096\}$ and output lengths $T \in \{128, 2048\}$ under the same companion-perturbation protocol. Blazedit and BurstGPT each contribute one four-request representative group with 64 generated tokens per decoded request. Rows is faster than CoRun in four of the six groups. At $W=4096$, Rows takes 9.176/141.141 s for 128/2048 output tokens, versus CoRun's 9.016/135.103 s; the phase breakdown reports this cost directly.}




\section{Discussion and Future Work}

{Operating Regimes: Rows vs. Pack. The choice between the two BlockLock primitives depends on the serving workload. BlockLock-Rows retains full $B$-request batched decoding and avoids inter-segment masking overhead ($O(BW^2)$ attention cost), making it ideal for throughput-oriented evaluation and RL rollout generation where all requests in a batch are decoded. Conversely, BlockLock-Pack is tailored for mixed-criticality auditing or canary verification where companion requests only require prefill processing (e.g., log-likelihood scoring, guardrail classification, or disaggregated KV pre-warming) while a single protected target requires full autoregressive decoding after sharing the $[1, BW]$ prefill. While the native PyTorch implementation of Pack materializes a dense $O(B^2W^2)$ segment mask, our vLLM integration replaces this dense mask with PagedAttention metadata, substantially reducing prefill overhead (Figure~\ref{fig:perf}).}

{Complementarity with Batch-Invariant Kernels. Table 4 summarizes the trade-offs across determinism mechanisms. Batch-invariant kernels enforce invariance across dynamically changing batch sizes at the cost of custom per-hardware implementations and a $>2\times$ latency penalty across all requests. 
{BlockLock instead enforces determinism within an eligible fixed deployment using stock framework kernels. Its measured overhead depends on the length class, workload, and backend; deployment validation is required on each hardware-model configuration.}}

{Limitations and Future Work. 
{Our evaluation covers dense models only; evaluating BlockLock on mixture-of-experts (MoE) models is left to future work.} First, BlockLock's determinism contract holds per deployment configuration $d$; changing GPU architectures (e.g., Ampere to Hopper) or tensor-parallel degrees alters underlying kernel reduction trees and remains outside our scope. Second, when companion lengths exhibit extreme variance within a coarse length class $W$, padding waste increases (as observed at $W=4096$ in Table 3). Future work includes integrating canonical geometry locking with chunked prefill and prefix caching, as well as extending deterministic execution plans to Mixture-of-Experts (MoE) token routing where capacity-factor dropping can induce cross-request interference.}

\section{Conclusions}

Current LLM serving does not give reproducible outputs, because the batch changes the order of the floating point sums. Existing fixes pay for reproducibility with custom kernels, verification work, or by giving up shared prefill. BlockLock instead pins the geometry that the protected request sees and keeps shared batched prefill, with two schemes: BlockLock-Rows uses fixed-width rows and BlockLock-Pack uses terminal-anchored packing with an isolated dummy segment. Under a stated per-deployment contract, the output of the protected request is bitwise identical for any companions. Across the eight-dataset benchmark, 
{at $B=4$ and 64 output tokens per request,} BlockLock-Rows adds only 0.5--0.8\% latency overhead over standard PyTorch padded batching while reducing vLLM Batch-Invariant (BI) execution time by 53.5--55.0\%. Extended capacity sweeps up to $W = 4096$ and $T = 2048$ further confirm that BlockLock-Rows outperforms sequential isolation across moderate context lengths, while BlockLock-Pack offers a low-latency alternative for single-target auditing when class padding is extreme. These results support canonical scheduling as an efficient way to protect a target while retaining batched prefill.

\clearpage

\bibliography{refs}
\bibliographystyle{mlsys2025}

\clearpage
\section*{AI Usage Disclosure}

Following the MLSys policy, we disclose the use of AI tools in the preparation of this paper, by category:
\begin{itemize}
\item \textbf{Ideation and experiment design:} a general-purpose LLM assistant was used to discuss design alternatives and evaluation plans; the authors made all the decisions.
\item \textbf{Literature search and reference checking:} a general-purpose LLM assistant was used to find and verify the cited references.
\item \textbf{Code and experiments:} a general-purpose LLM assistant was used for code review and for help with the experimental infrastructure; the authors verified the results.
\item \textbf{Figures and writing:} generative AI tools were used to produce the schematic figures and for writing and editing support; the authors revised the final text and figures.
\end{itemize}
Large language models are also the object of study of this paper: they are the models evaluated in the experiments. The authors designed the work, verified the results, and take full responsibility for the content of this paper.

\end{document}


\twocolumn[
\mlsystitle{BlockLock: Reproducible LLM Inference with Fixed Batch Geometry}

\begin{mlsysauthorlist}
\mlsysauthor{Xiuyuan Qi}{uestc}
\mlsysauthor{Pedro Reviriego}{upm}
\mlsysauthor{Javier Conde}{upm}
\mlsysauthor{Shanshan Liu}{uestc}
\end{mlsysauthorlist}
\mlsysaffiliation{uestc}{University of Electronic Science and Technology of China, Chengdu, Sichuan, China}
\mlsysaffiliation{upm}{Universidad Politécnica de Madrid, Madrid, Community of Madrid, Spain}

\mlsyscorrespondingauthor{Pedro Reviriego}{pedro.reviriego@upm.es}
\mlsyscorrespondingauthor{Shanshan Liu}{ssliu@uestc.edu.cn}
\vskip 0.3in
]

\appendix

\section{Formal Design and Determinism Contract}
\label{app:formal}

BlockLock coordinates execution geometry, token layout, backend eligibility, and distributed collective execution to preserve a protected target request's numerical path through prefill and decode. This appendix formalizes the canonical execution groups, the two prefill primitives (\texttt{BlockLock-Rows} and \texttt{BlockLock-Pack}), the request-local KV cache extraction for the packed path, and the deployment-level determinism contract along with its proof.

\subsection{Canonical Execution Groups and Mixed-Criticality Co-Batching}
\label{app:groups}

The serving workload comprises a protected target request $x$ requiring bitwise reproducibility and a set of ordinary companion requests $F$ that share accelerator compute. Our controlled companion-perturbation protocol varies the companion set $F$ around a fixed target $x$, realizing mixed-criticality co-batching at the execution-group level.

Let $L_i$ denote the tokenized prompt length of request $i$, and let $L$ denote the capacity-alignment block granularity (e.g., $L = 64$ or $128$ tokens, matching Section~3.1 of the main text). Initial length buckets are indexed by $k_i = \lceil L_i / L \rceil$. Adjacent buckets may be merged into deterministic length classes with a fixed class width $W_i = m_i L \geq L_i$ ($m_i \in \mathbb{Z}^+$). The length-class mapping is frozen across all compared executions within a deployment. For a logical batch of $B$ co-scheduled requests belonging to a class of width $W$, BlockLock reserves a total capacity of
\begin{equation}
P = B W
\end{equation}
prompt token positions. Each group contains the protected target $x$ and $B - 1$ eligible companions $F = \{f_1, \ldots, f_{B-1}\}$. Across repetitions, a seeded schedule perturbs both the token contents and exact lengths $L_f \le W$ of the companions while holding the target $x$, batch size $B$, class width $W$, and canonical layout strictly fixed.

\paragraph{Distributed execution plan ($TP=2$).}
The same logical plan governs both single-GPU execution and multi-GPU tensor parallelism ($TP = 2$) across two NVIDIA A6000 GPUs. The distributed runner initializes a \texttt{torch.distributed} NCCL process group with one rank per GPU and shards the model via the native Transformers tensor-parallel interface (\texttt{tp\_plan="auto"} with \texttt{tp\_size=2}). Rank~0 constructs and broadcasts the canonical batch plan (token IDs, padding/segment masks, and position IDs) to all ranks. Before executing the forward pass, all ranks verify via hash checks that their received input plans and model sharding topologies match identically. Consequently, all column- and row-parallel linear layers execute an identical sequence of collective communications across both prefill and decode phases.

Rank assignment, tensor-sharding boundaries, software versions, and NCCL collective configurations jointly define the distributed deployment environment \citep{pytorch_distributed,nvidia_nccl_env}. Because NCCL dynamically selects communication rings/trees and protocols based on tensor shapes and topology, our contract requires that the chosen collective algorithms be repeat-stable under fixed tensor shapes (Condition~C2).

\subsection{Two Canonical Prefill Primitives: Row Locking and Packed Locking}
\label{app:primitives}

Both BlockLock primitives reserve $P = BW$ prompt positions and enforce the same backend eligibility policy, but expose distinct attention-masking and decode-state trade-offs tailored to different serving regimes.

\paragraph{Primitive 1: Fixed-width rows (\texttt{BlockLock-Rows}).}
The batch is structured as a 2D tensor of shape $[B, W]$. Each request occupies a dedicated row of fixed width $W$, where left-padding aligns real tokens to the right boundary of the row and a 2D padding mask isolates valid tokens from padding. Because the target request $x$ occupies a fixed row index with invariant logical positions and row width $W$, row-wise operators and attention reductions observe the exact same tensor geometry regardless of how companion lengths vary in other rows. A single batched prefill invocation processes all $B$ rows simultaneously, and decoding proceeds via standard $B$-request batched KV caching.

For dense full attention (omitting head and hidden-dimension factors), attention score computation requires $O(B W^2)$ work. Crucially, because requests are separated along the batch dimension, \texttt{BlockLock-Rows} requires no dense 2D inter-segment causal mask, giving it a significant complexity advantage over \texttt{BlockLock-Pack} as $B$ scales. Moreover, because every row in $[B, W]$ has fixed width $W$, any row whose request is held fixed enjoys the same geometric invariance as the audited target row.

\paragraph{Primitive 2: Terminal-anchored capacity packing (\texttt{BlockLock-Pack}).}
In \texttt{BlockLock-Pack}, the $B - 1$ companions, an isolated dummy filler segment, and the protected target $x$ are concatenated along the sequence dimension into a single packed block of shape $[1, P]$. Let $R = L_x + \sum_{f \in F} L_f$ denote the total number of real prompt tokens in the group. The remaining unused capacity is absorbed by a dummy segment of length
\begin{equation}
D = P - R = B W - L_x - \sum_{f \in F} L_f \geq 0.
\end{equation}
By placing the companions first, the $D$-token dummy segment second, and the protected target $x$ strictly last, $x$ is anchored at the fixed terminal physical interval
\begin{equation}
I_x = [P - L_x, P).
\end{equation}
Consequently, both the start offset $P - L_x$ and end offset $P$ of the target remain completely invariant to fluctuations in companion lengths $\sum_{f \in F} L_f$. For instance, with $W = 192$ and $B = 2$ ($P = 384$), a companion of length $180$ and a target of length $192$ induce a $D = 12$-token dummy segment, leaving $I_x = [192, 384)$ unchanged.

Within the packed sequence, each real segment $i$ is assigned reset local position IDs $0, 1, \ldots, L_i - 1$ (for RoPE embeddings). Let $s(u)$ denote the segment index of physical token position $u \in [0, P)$, where the dummy segment is assigned its own isolated segment identity. In the native PyTorch implementation, logical isolation is enforced via an additive segment-causal attention mask $A \in \mathbb{R}^{1 \times 1 \times P \times P}$:
\begin{equation}
A_{uv} =
\begin{cases}
0, & \text{if } s(u) = s(v) \text{ and } v \leq u, \\
m_{\mathrm{dtype}}, & \text{otherwise},
\end{cases}
\end{equation}
where $m_{\mathrm{dtype}} = \texttt{torch.finfo(dtype).min}$. This mask guarantees strictly segment-local causal visibility. A single forward pass over $[1, P]$ computes the packed KV cache and yields the target's initial next-token logits at position $P - 1$. In native PyTorch, materializing $A_{uv}$ requires $O(P^2) = O(B^2 W^2)$ memory and attention score work; in our vLLM integration, segment isolation is instead handled via PagedAttention block tables and segment metadata without materializing a dense $[P, P]$ matrix.

\begin{algorithm}[t]
\caption{\texttt{BlockLock-Pack}: Terminal-Anchored Packed Prefill and Target-Local Decode}
\label{alg:scheme2}
\begin{algorithmic}[1]
\REQUIRE Target prompt $x$ (length $L_x$), companions $F$, class width $W$, batch size $B$, decode horizon $T$
\STATE $P \leftarrow B W$; \quad $D \leftarrow P - L_x - \sum_{f \in F} L_f$ \quad (\textbf{require} $D \geq 0$)
\STATE Pack sequence of shape $[1, P]$ by concatenating $F$, an isolated $D$-token dummy segment (if $D > 0$), and $x$ at interval $I_x = [P - L_x, P)$
\STATE Construct segment-reset local position IDs and segment-causal mask $A_{uv}$
\STATE $(\ell_0, K_{\mathrm{pack}}) \leftarrow \mathrm{Prefill}(\text{packed tokens}, \text{positions}, A_{uv})$
\STATE $K_x \leftarrow \mathrm{ExtractTargetKV}(K_{\mathrm{pack}}, I_x, L_x)$
\FOR{$t = 0$ \textbf{to} $T - 1$}
    \STATE $y_t \leftarrow \mathrm{Greedy}(\ell_t)$; \quad record token $y_t$ and raw logits $\ell_t$
    \IF{$t < T - 1$}
        \STATE $(\ell_{t+1}, K_x) \leftarrow \mathrm{Decode}(y_t, K_x, \text{pos} = L_x + t)$
    \ENDIF
\ENDFOR
\STATE 
\textcolor{black}{\textbf{return}} Target token sequence $\{y_t\}_{t=0}^{T-1}$, raw-logit trace, and CUDA phase timings
\end{algorithmic}
\end{algorithm}

\subsection{Request-Local KV Extraction and Decode Handoff}
\label{app:kv}

Following the shared prefill in \texttt{BlockLock-Pack} (Algorithm~\ref{alg:scheme2}), BlockLock constructs a standalone target-local KV cache (\texttt{DynamicCache} in Transformers or target block table in vLLM) by extracting only $x$'s state layer by layer:
\begin{itemize}
\item \textbf{Full-attention layers:} The extractor slices and retains the terminal $L_x$ key and value states corresponding to $I_x = [P - L_x, P)$.
\item \textbf{Sliding-window attention layers:} For models employing hybrid full/sliding-window attention (such as Gemma-3-4B/12B-IT), the extractor retains the visible target suffix within the sliding window and restores the cache's logical cumulative sequence length metadata (`\_seen\_tokens`) to $L_x$.
\end{itemize}
This handoff guarantees that every KV entry visible to $x$'s attention window is preserved bit-for-bit without value conversion. Subsequent decode steps execute with input shape $[1, 1]$, target-local position IDs $L_x + t$, and the isolated cache $K_x$. By decoding $x$ in isolation rather than batch-decoding companions with varying unpadded lengths $L_f$, \texttt{BlockLock-Pack} ensures that the decode phase also strictly satisfies canonical geometry (Condition~C1).

\subsection{Configuration-Level Determinism Contract and Proof}
\label{app:contract}

Let $x$ denote the protected target prompt, $\Theta$ the frozen model weights, and $d$ the deployment configuration tuple:
\begin{equation}
\begin{aligned}
d = (&\text{Hardware},\text{Software},\text{Precision},\\
     &\text{BackendPolicy},\text{ClassMap},B,L,\\
     &\text{Primitive},\text{TPConfig}).
\end{aligned}
\end{equation}
Each distinct GPU architecture (A6000 Ampere, A100 Ampere, H100 Hopper, or $2\times$A6000 $TP=2$) constitutes a distinct deployment $d$. Let $\pi$ denote the companion composition and scheduling instance, $S(x, d)$ the canonical execution plan assigned to $x$, and $L_t(x; \Theta, d, \pi)$ the raw full-vocabulary logit vector produced for target $x$ at generation step $t$ prior to argmax selection.

\paragraph{Definition 1 (Per-Deployment Bitwise Reproducibility Contract).}
\emph{For any two companion schedules $\pi_1$ and $\pi_2$ compatible with the canonical plan $S(x, d)$ under a fixed deployment $d$, the raw logit vectors of target $x$ are bitwise identical at every generation step $t$:}
\begin{equation}
\mathrm{bits}\!\left(L_t(x; \Theta, d, \pi_1)\right) = \mathrm{bits}\!\left(L_t(x; \Theta, d, \pi_2)\right), \quad \forall t \geq 0.
\end{equation}

The contract relies on three sufficient conditions (stated in Section~3.4 of the main text):
{\renewcommand{\labelenumi}{\textbf{C\arabic{enumi}.}}
\begin{enumerate}
\item \textbf{Canonical Geometry and Logical Isolation:} The target's tensor dimensions, row/terminal physical offsets, local position IDs, and causal attention masks remain invariant across $\pi_1$ and $\pi_2$, preventing companion tokens from contributing to target attention values. In $TP=2$, shard boundaries and collective call sequences are likewise invariant.
\item \textbf{Repeat-Stable Operator and Collective Arithmetic:} Under a fixed tensor shape and alignment, all enabled kernel dispatches and distributed NCCL collectives are deterministic across repeated invocations, and companion values in masked/other-row positions do not alter the floating-point reduction tree of the target's positions.
\item \textbf{Faithful Cache-State Handoff:} The transition from prefill to decode preserves the exact bitwise KV tensors, sliding-window suffixes, and position metadata of the target without numerical cast or truncation of visible states.
\end{enumerate}}

\paragraph{Proposition 1 (Conditional Deterministic State Preservation).}
\emph{Under a fixed deployment $d$ and canonical plan $S(x, d)$ satisfying Conditions C1--C3, deterministic greedy decoding produces bitwise-identical raw logits $L_t$, identical generated token IDs $y_t$, and identical target KV states $K_x^{(t)}$ across all compatible companion schedules $\pi$.}

\paragraph{Proof.}
We proceed by induction on the generation step $t$.
\emph{Base case ($t = 0$, Prefill):} By Condition~C1, target $x$ occupies the exact same row in $[B, W]$ (\texttt{Rows}) or terminal interval $I_x = [P - L_x, P)$ in $[1, P]$ (\texttt{Pack}) with identical local positions $0, \ldots, L_x - 1$ and causal masking under both $\pi_1$ and $\pi_2$. By Condition~C2, every linear projection, normalization, attention kernel, and TP collective executes the exact same reduction tree over $x$'s isolated representation. Thus, the prefill step produces bitwise-identical initial target logits $\mathrm{bits}(L_0(\pi_1)) = \mathrm{bits}(L_0(\pi_2))$ and identical visible target KV entries. By Condition~C3, the decode handoff yields bitwise-identical target KV caches $K_x^{(0)}(\pi_1) = K_x^{(0)}(\pi_2)$.

\emph{Inductive step ($t \to t + 1$, Decode):} Suppose that at step $t$, both executions share bitwise-identical token histories $y_{0:t-1}$, raw logits $L_t$, and target KV states $K_x^{(t)}$. Deterministic greedy selection ($\mathrm{argmax}$ with fixed tie-breaking) selects the exact same token $y_t(\pi_1) = y_t(\pi_2)$. At step $t + 1$, the decode invocation receives identical input token $y_t$, position ID $L_x + t$, cache state $K_x^{(t)}$, and step geometry ($[B, 1]$ with row-independent reductions in \texttt{Rows}, or $[1, 1]$ in \texttt{Pack}). By Conditions~C1 and C2, the resulting updated KV cache $K_x^{(t+1)}$ and next-step raw logits $L_{t+1}$ are bitwise identical: $\mathrm{bits}(L_{t+1}(\pi_1)) = \mathrm{bits}(L_{t+1}(\pi_2))$. By induction, the claim holds for all $t \geq 0$.

---

\section{Experimental Details and Protocol}
\label{app:exp}

This section provides the details on the experimental setup. The code and results are available at the following repository: \emph{Omitted for double-blind review}.

\subsection{Hardware Platforms and Software Stack}
Our evaluation spans three NVIDIA GPU platforms across two architectures:
\begin{itemize}
\item \textbf{NVIDIA RTX A6000 (48~GB, Ampere):} Single-GPU evaluation and dual-GPU tensor-parallel ($TP = 2$) evaluation via NVLink/PCIe with NCCL.
\item \textbf{NVIDIA A100 (80~GB, Ampere):} Single-GPU evaluation.
\item \textbf{NVIDIA H100 NVL (Hopper):} Single-GPU evaluation (including MIG \texttt{3g.47gb} slices in backend audits and full-device runs).
\end{itemize}

The software stack uses Python~3.11, PyTorch~2.11.0+cu128, HuggingFace Transformers~5.13.0, vLLM (with native Batch-Invariant kernel support enabled for \texttt{vLLM BI}), CUDA~12.8, and cuDNN~9.19. To satisfy Condition~C2, our restricted backend policy sets \texttt{torch.backends.cuda.matmul.allow\_tf32 = False}, \texttt{torch.backends.cudnn.allow\_tf32 = False}, and disables cuDNN SDPA (\texttt{enable\_cudnn\_sdp(False)}), while retaining FlashAttention-2, Memory-Efficient SDPA, and Math SDPA according to hardware/dtype eligibility \citep{pytorch_sdpa,pytorch_reproducibility}. All models run in \texttt{eval()} mode with \texttt{bfloat16} (\texttt{BF16}) weights and activations and a fixed seed of $0$.

\subsection{Evaluated Models}
We evaluate five open-weight 
\textcolor{black}{dense} instruction-tuned and base models spanning three model families (including dense standard attention, Grouped-Query Attention, and Gemma-3's interleaved local sliding-window / global full attention):
\begin{itemize}
\item \textbf{Qwen3-8B:} \url{https://huggingface.co/Qwen/Qwen3-8B}
\item \textbf{Qwen3-14B:} \url{https://huggingface.co/Qwen/Qwen3-14B}
\item \textbf{Llama-3.1-8B-Instruct:} \url{https://huggingface.co/meta-llama/Llama-3.1-8B-Instruct}
\item \textbf{Gemma-3-4B-IT:} \url{https://huggingface.co/google/gemma-3-4b-it}
\item \textbf{Gemma-3-12B-IT:} \url{https://huggingface.co/google/gemma-3-12b-it}
\end{itemize}

\subsection{Datasets and Workloads}
\begin{itemize}
\item \textbf{Eight-Dataset Core Benchmark:} Comprises seven suites from \textbf{tinyBenchmarks} (\url{https://huggingface.co/tinyBenchmarks}) \citep{maiapolo2024tinybenchmarks}---MMLU, AI2 ARC, HellaSwag, GSM8K, TruthfulQA, WinoGrande, and AlpacaEval---plus multi-function tool-use queries from the \textbf{Berkeley Function Calling Leaderboard (BFCL v3)} (\url{https://gorilla.cs.berkeley.edu/}) \citep{patil2025bfcl}. BFCL queries are evaluated in both standard prompt-rendering mode and native tool-calling chat-template mode where supported by the tokenizer. (The 188-query BFCL study cited in Section~1 is an independent motivational audit.)
\item \textbf{Long-Context and Real-World Trace Workloads (Table~3 of main text):} Evaluates synthetic stress tests with class widths $W \in \{1024, 4096\}$ and output lengths $T \in \{128, 2048\}$ alongside real-world variable-length prompt traces sampled from \textbf{Blazedit} and \textbf{BurstGPT} ($T = 64$ generated tokens per decoded request, $B = 4$). 
\textcolor{black}{Blazedit and BurstGPT each supply one representative four-request group; these are controlled tests, not full arrival-trace replays.}
\item \textbf{Implementation of vLLM Rows:} In our vLLM adapter, BlockLock-Rows explicitly enforces canonical prefill geometry by constructing a flattened input tensor of exactly $BW$ tokens, prepending $W - L_i$ left-padding tokens before request $i$ such that: 
$$\texttt{query\_start\_loc} = [0, W, 2W, \ldots, BW].$$    

Inserting physical query padding tokens is necessary because merely shifting sequence offsets in a ragged tensor would still expose the projection, normalization, and FFN layers to companion-dependent total token counts $\sum_i L_i$. Our vLLM integration decouples prefill query padding from decode KV storage. Zero KV Cache Pollution ($L=SlotMapping = -1$): All $W - L_i$ padding positions are assigned $SlotMapping = -1$, preventing them from allocating or writing to PagedAttention KV blocks.
Exact Causal Alignment: Each request retains its true context length $seq_lens[i] = Li$ and local RoPE positions $0, \ldots, L_i - 1$  , while FlashAttention's bottom-right causal alignment isolates leading padding queries from the valid prefix. The adapter also pins attention bounds, enforces a single attention split (disabling non-deterministic Split-KV), and replays captured CUDA graphs.
\end{itemize}

\subsection{Companion-Perturbation Protocol and Agreement Metrics}
Across batch sizes $B \in \{1, 2, 4, 8, 16\}$, each target question in a dataset is evaluated over $r = 10$ independent repetitions. In repetition $k \in \{1, \ldots, 10\}$, the target prompt $x$ is co-scheduled with a distinct, randomly sampled tuple of $B - 1$ companion prompts drawn from the same length class. This yields $\binom{10}{2} = 45$ pairwise cross-companion comparisons per (platform, model, dataset, target, $B$) condition, totaling \textbf{1,316} single-GPU conditions and \textbf{20} $TP = 2$ conditions (60,120 pairwise comparisons per scheme).

\paragraph{Generated-Sequence Agreement ($A_{\mathrm{seq}}$).}
Suppose $r$ repetitions of a fixed target partition into distinct generated token-ID sequences with multiplicities $n_1, n_2, \ldots, n_m$ ($\sum_{j=1}^m n_j = r$). The within-condition pairwise sequence agreement is
\begin{equation}
A_{\mathrm{seq}} = \frac{\sum_{j=1}^m \binom{n_j}{2}}{\binom{r}{2}} \in [0, 1].
\end{equation}
$A_{\mathrm{seq}} = 100\%$ if and only if all $r = 10$ companion-perturbed runs produce the exact same token sequence ($m = 1, n_1 = 10$).

\paragraph{Top-20 Logit-Trace Agreement ($A_{\mathrm{top20}}$) and Full-Vocabulary Hash Audit.}
Even when greedy argmax tokens agree, subtle floating-point drift can alter underlying logits. At every generation step $t$, we record the top-20 candidate token IDs and their exact 16-bit \texttt{BF16} raw logit bit patterns. Two runs agree in Top-20 trace if and only if both the ordered top-20 candidate IDs and their raw logit bits match identically at every step $t$. In our vLLM and PyTorch 8-dataset benchmarks (80 conditions per method), we additionally compute SHA-256 hashes over the full-vocabulary raw logit tensor at every step to verify strict bitwise identity across the entire vocabulary.

\subsection{Phase Timing Instrumentation and Dual-Engine Scope}
Prefill and decode latencies are bracketed using hardware \texttt{torch.cuda.Event(enable\_timing=True)} markers synchronized around the respective model forward calls. Because our experiments evaluate deterministic batch execution mechanics rather than network queuing, we report exact prefill and decode phase durations rather than arrival-queue TTFT or TPOT.
\begin{itemize}
\item \textbf{Equal-Delivered-Work Benchmarks (\texttt{STD}, \texttt{vLLM BI}, \texttt{Padded}, \texttt{CoRun-style}, \texttt{BlockLock-Rows}):}

All $B$ requests in the batch are prefilled and decoded for $T$ tokens. The standard agreement and timing suite, including all panels in Section~\ref{app:detailed_results}, uses $T=500$; the small-scale dual-engine (vLLM vs. PyTorch) benchmark uses its separately specified decode horizon.
\item \textbf{Single-Target Audit Benchmarks (\texttt{BlockLock-Pack}$^*$):} All $B$ requests share the $[1, BW]$ prefill, after which only the protected target $x$ is decoded. The results in Section~\ref{app:detailed_results} use $T=500$ target tokens; the small-scale dual-engine audit uses $T=64$.
\end{itemize}
For the small-scale dual-engine benchmark with $B=4$ and $T=64$, actual output throughput (tokens/s) is computed as $\sum_{r} N_r / (\sum_{r} T_r / 1000)$, where $N_r$ is the number of generated tokens ($256$ for full-batch methods; $64$ for \texttt{Pack}$^*$) and $T_r$ is the stacked prefill-plus-decode duration in milliseconds. Memory comparisons enforce identical logit-capture modes across methods to prevent diagnostic trace buffers from skewing peak GPU allocation measurements.

\subsection{Condition Breakdown and Backend Eligibility Filtering}

\begin{table*}[!htbp]
\centering
\caption{Pair-weighted generated-sequence agreement (\%) and Top-20 raw-logit agreement (\%) across the evaluated single-GPU and $TP=2$ configurations with T = 500 generated output tokens per request ($B \in \{1, 2, 4, 8, 16\}$, 10 repetitions per condition). 
\textcolor{black}{The single-GPU subtotal and the combined total are distinct populations.} \texttt{CoRun}, \texttt{BlockLock-Rows}, and \texttt{BlockLock-Pack} achieve \textbf{100.0\%} across all 1,336 conditions.}
\label{tab:normal-agreement}
\footnotesize
\setlength{\tabcolsep}{4pt}
\begin{tabular}{llrrr}
\toprule
Platform & Model & \begin{tabular}[c]{@{}r@{}}Padded Seq.\\ Agree. (\%)\end{tabular} & \begin{tabular}[c]{@{}r@{}}BlockLock\\ Agree. (\%)\end{tabular} & \begin{tabular}[c]{@{}r@{}}Evaluated\\ Conditions\end{tabular} \\
\midrule
A100 & Qwen3-14B & 57.4 & 100.0 & 135 \\
A100 & Qwen3-8B & 56.6 & 100.0 & 159 \\
A6000 & Qwen3-14B & 55.8 & 100.0 & 105 \\
A6000 & Qwen3-8B & 62.9 & 100.0 & 160 \\
A6000 & Gemma-3-4B-IT & 40.4 & 100.0 & 130 \\
A6000 & Llama-3.1-8B & 44.9 & 100.0 & 160 \\
H100 & Qwen3-14B & 55.8 & 100.0 & 160 \\
H100 & Qwen3-8B & 57.5 & 100.0 & 160 \\
H100 & Llama-3.1-8B & 51.1 & 100.0 & 147 \\
\midrule
\multicolumn{2}{l}{\textbf{Single-GPU Subtotal / Mean}} & \textbf{52.59} & \textbf{100.0} & \textbf{1,316} \\
\multicolumn{2}{l}{\textbf{(Top-20 Logit Agreement)}} & \textbf{(35.93)} & \textbf{(100.0)} & \textbf{1,316} \\
\midrule
$2\times$A6000 ($TP=2$) & Gemma-3-4B-IT & 59.4 & 100.0 & 20 \\
\midrule
\multicolumn{2}{l}{\textbf{Total Evaluated Conditions}} & \textbf{52.69} & \textbf{100.0} & \textbf{1,336} \\
\bottomrule
\end{tabular}
\end{table*}

Table~\ref{tab:normal-agreement} details the per-configuration breakdown of the 1,316 single-GPU conditions and 20 dual-GPU ($TP=2$) conditions summarized in Section~4.2 of the main text. Across all 1,316 single-GPU conditions, standard padded batching (\texttt{STD}) achieves only \textbf{52.59\%} pair-weighted generated-sequence agreement and \textbf{35.93\%} Top-20 raw-logit agreement, whereas \texttt{CoRun}, \texttt{BlockLock-Rows}, and \texttt{BlockLock-Pack} achieve \textbf{100.0\%} on both metrics.

---

\section{Single-Request ($B = 1$) Backend Audit and Kernel Intervention}
\label{app:backend_audit}

A critical finding of our study (Section~2.3) is that LLM inference can fail to be reproducible even at batch size $B = 1$ with fixed seeds and greedy decoding if the underlying kernel dispatch violates repeat stability (Condition~C2).

To isolate backend-induced nondeterminism from batch-composition effects, we conducted a controlled $B = 1$ audit on an NVIDIA H100 NVL GPU (MIG \texttt{3g.47gb} instance) running Qwen3-8B in \texttt{BF16} on a fixed 117-token tinyMMLU prompt over 320 greedy decoding steps. Even though the logical input and tensor geometry were held completely static across repetitions, PyTorch's default SDPA backend dispatcher selected cuDNN's fused attention kernel, 
\textcolor{black}{on the divergent execution path. This identifies the dispatched backend, but does not establish which internal reduction or workspace operation caused the variation.}

\begin{table*}[!htbp]
\centering
\caption{Single-request ($B = 1$) repeat-stability audit on NVIDIA H100 (Qwen3-8B, \texttt{BF16}, 117-token prompt, 320 greedy steps). Counts indicate unique generated sequences / total repeated runs.}
\label{tab:b1_audit}
\footnotesize
\setlength{\tabcolsep}{5pt}
\begin{tabular}{lccc}
\toprule
Execution Path & \begin{tabular}[c]{@{}c@{}}Default Backend\\ (cuDNN SDPA + TF32)\end{tabular} & \begin{tabular}[c]{@{}c@{}}Restricted Policy\\ (No cuDNN SDPA / TF32)\end{tabular} & \begin{tabular}[c]{@{}c@{}}Raw-Logit\\ Bit Match\end{tabular} \\
\midrule
CoRun($B=1$) & 2 / 4 unique (Divergent) & 1 / 4 unique (Identical) & 100\% \\
Padded ($B=1$) & 3 / 3 unique (Divergent) & 1 / 3 unique (Identical) & 100\% \\
BlockLock-Rows ($B=1$) & 2 / 3 unique (Divergent) & 1 / 3 unique (Identical) & 100\% \\
\bottomrule
\end{tabular}
\end{table*}

As shown in Table~\ref{tab:b1_audit}, under the default backend policy, repeating the exact same $B = 1$ execution yielded 2 unique sequences out of 4 runs for \texttt{CoRun}, 3 out of 3 distinct sequences for \texttt{Padded}, and 2 out of 3 for \texttt{BlockLock-Rows}. Kernel-level profiling via NVIDIA Nsight Systems confirmed that cuDNN SDPA was invoked on the divergent path. When our restricted backend policy disables cuDNN SDPA and TF32 
\textcolor{black}{(restricting the eligible attention backend set)}, all repeated runs collapsed to a single unique sequence ($1/4, 1/3, 1/3$) with \textbf{100\% bitwise raw-logit identity} across all 320 steps. This empirical evidence motivates Condition~C2 of our contract: canonical geometry (Condition~C1) must be paired with a verified repeat-stable backend policy.

---

\section{Deployment Validation and Reproducibility Checklist}
\label{app:checklist}

\textcolor{black}{To validate the conditions required for end-to-end bitwise reproducibility in production or evaluation pipelines,} each BlockLock deployment enforces the following automated validation checklist prior to serving:
\begin{enumerate}
\item \textbf{Decoding and Seed Lock:} Enforce deterministic greedy selection (or request-bound RNG state for sampling) with fixed tie-breaking and stopping criteria.
\item \textbf{Backend Eligibility Gate (Condition~C2):} Disable non-deterministic kernel dispatches (\texttt{cudnn\_sdp = False}, \texttt{allow\_tf32 = False}) 
\textcolor{black}{and use a $B = 1$ repeat check as an initial canary. Validation must also cover the deployed model, precision, prefill/decode shapes, length classes, and TP configuration; passing the canary alone does not establish Condition C2 for other dispatches.}
\item \textbf{Environment Pinning:} Record exact hardware SKU, driver/CUDA/cuDNN versions, and framework releases (PyTorch, Transformers, vLLM).
\item \textbf{Canonical Geometry Lock (Condition~C1):} Freeze the alignment granularity $L$, length-class width mapping $W$, batch size $B$, and primitive layout (\texttt{Rows} $[B, W]$ or \texttt{Pack} $[1, BW]$ with terminal interval $I_x = [P - L_x, P)$).
\item \textbf{Multi-GPU Collective Verification ($TP > 1$):} Verify via pre-inference rank broadcasts and post-inference SHA-256 logit-hash comparisons that all tensor-parallel ranks execute identical plans and repeat-stable NCCL reductions.
\item \textbf{Audit Logging:} Optionally persist generated token IDs, Top-20 raw-logit traces (or full-vocabulary SHA-256 hashes), and CUDA-event phase intervals for post-hoc verification.
\end{enumerate}
Any modification to hardware topology, driver/library versions, or batch-class geometry constitutes a new deployment $d$ and triggers re-validation of the checklist.

\clearpage
\section{Detailed Experimental Results}
\label{app:detailed_results}

This section presents generated-sequence agreement, Top-20 raw-logit agreement where available, and prefill/decode timings for individual model--GPU configurations across the eight datasets. All experiments shown in this section use \texttt{BF16}, batch sizes $B \in \{1,2,4,8,16\}$, ten companion-perturbed repetitions per condition, and a decode horizon of $T=500$ tokens per decoded request. Agreement metrics follow Section~\ref{app:exp}. The target-question identifier and input mode are retained in each panel's original title where provided.

\paragraph{Method-name correspondence.}
The original plot labels \texttt{Normal}, \texttt{CoRun}, \texttt{Scheme 1}, and \texttt{Scheme 2} correspond to \texttt{STD}, \texttt{CoRun}, \texttt{BlockLock-Rows}, and \texttt{BlockLock-Pack}, respectively, in the text. In particular, \texttt{Normal} denotes standard padded batching, \texttt{Scheme 1} denotes fixed-width row locking, and \texttt{Scheme 2} denotes terminal-anchored capacity packing with target-local decode. We use the manuscript names below and retain the original labels inside the plots.

\paragraph{Reading the panels.}
Colors indicate batch size. Generated-sequence and Top-20 agreement are pairwise percentages across repeated runs. Timing bars stack the median prefill and median decode CUDA durations, with solid and pale fills identifying the two phases. Blank cells indicate unavailable or incomplete measurements. Under \texttt{BlockLock-Pack}, shared prefill is followed by decoding the protected target; the other methods decode all requests in the batch. All decoded requests in these panels use the same $500$-token horizon. Top-20 panels are included for configurations with an available export.

\newcommand{\ExpResultPanel}[2]{%
  \includegraphics[width=\textwidth]{#1}\par
  {\footnotesize #2\par}\medskip
}

\begin{figure*}[p]
\subsection{NVIDIA RTX A6000}
\label{app:results-a6000-5}

\centering
\ExpResultPanel{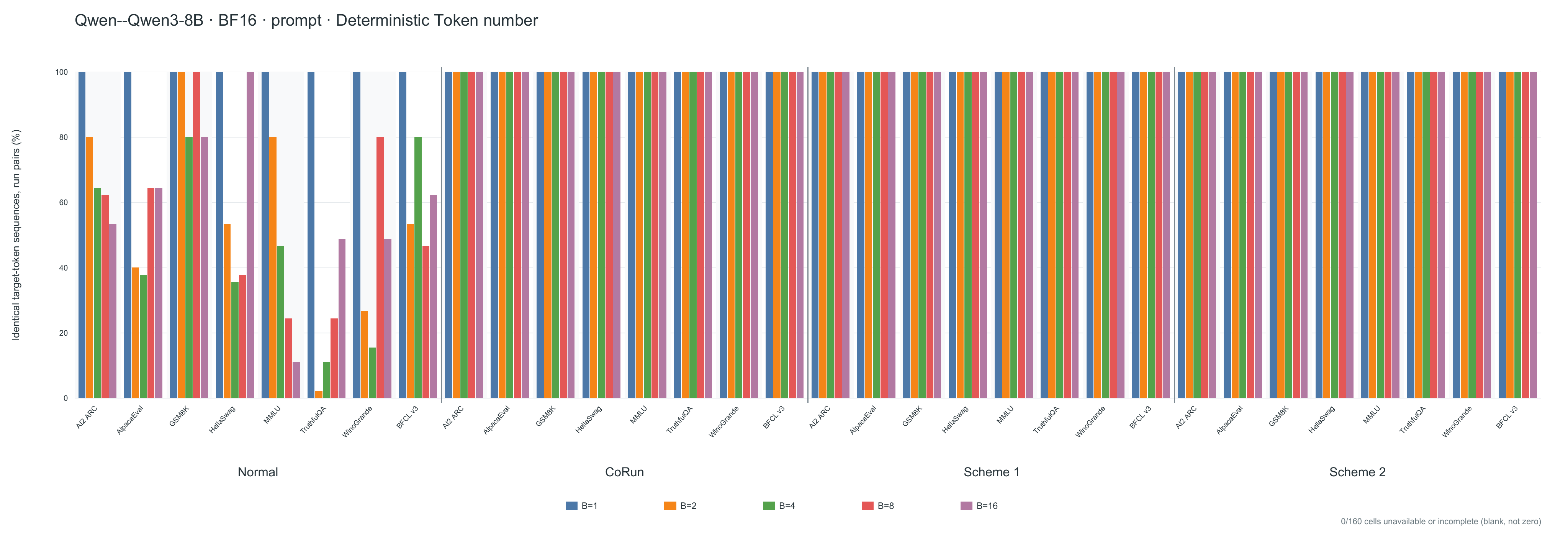}{(a) Generated-sequence agreement.}
\ExpResultPanel{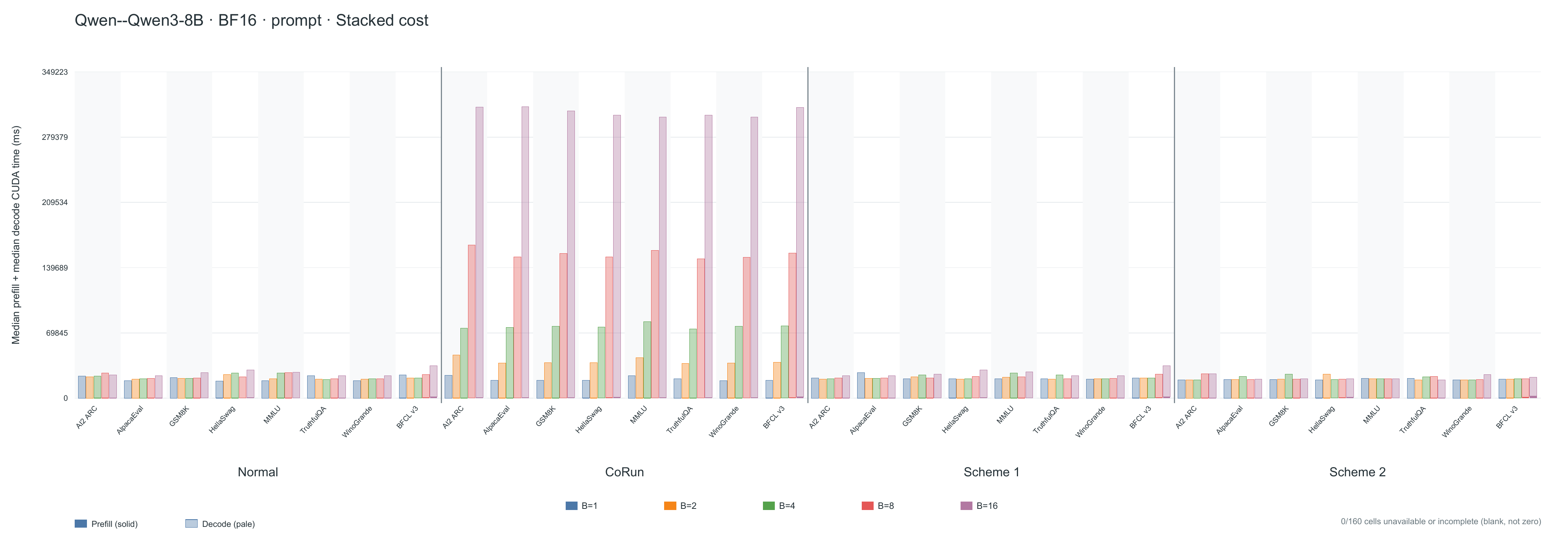}{(b) Prefill and decode CUDA time.}
\caption{Qwen3-8B on NVIDIA RTX A6000 (\texttt{BF16}, \texttt{prompt}, $T=500$): (a) generated-sequence agreement; (b) prefill and decode cuda time. Method labels follow the correspondence defined at the start of Section~\ref{app:detailed_results}.}
\label{fig:exp-a6000-5-qwen3-8b-prompt}
\end{figure*}

\begin{figure*}[p]
\centering
\ExpResultPanel{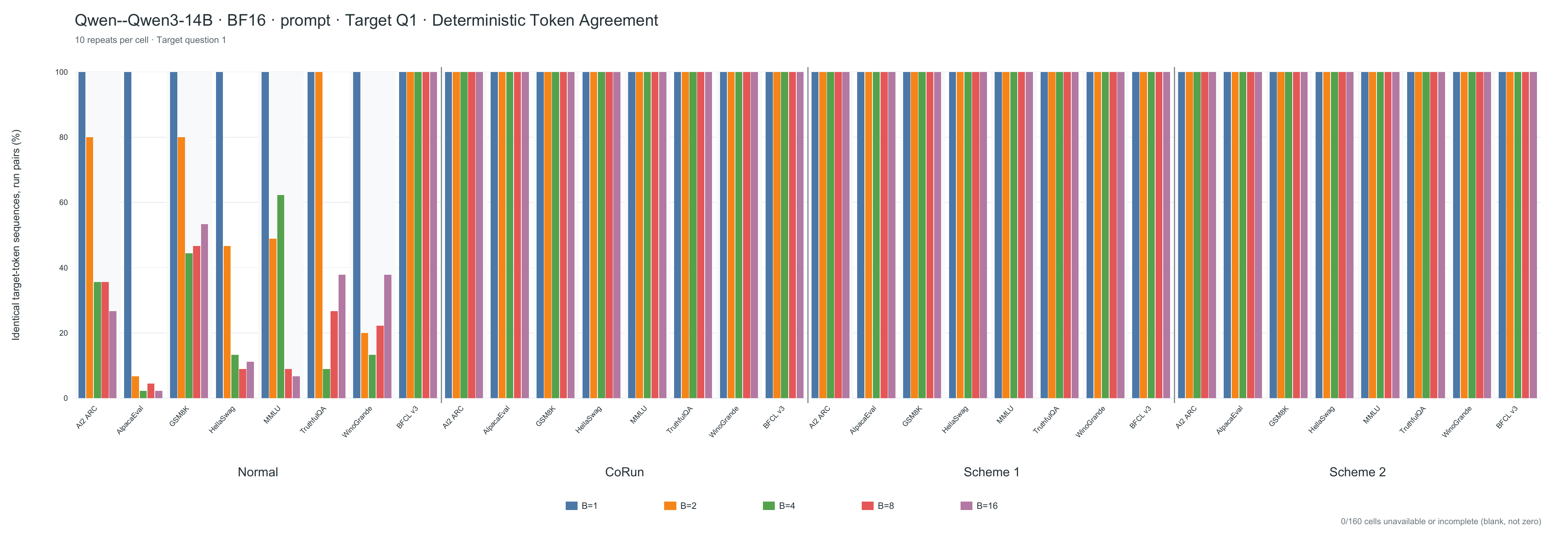}{(a) Generated-sequence agreement.}
\ExpResultPanel{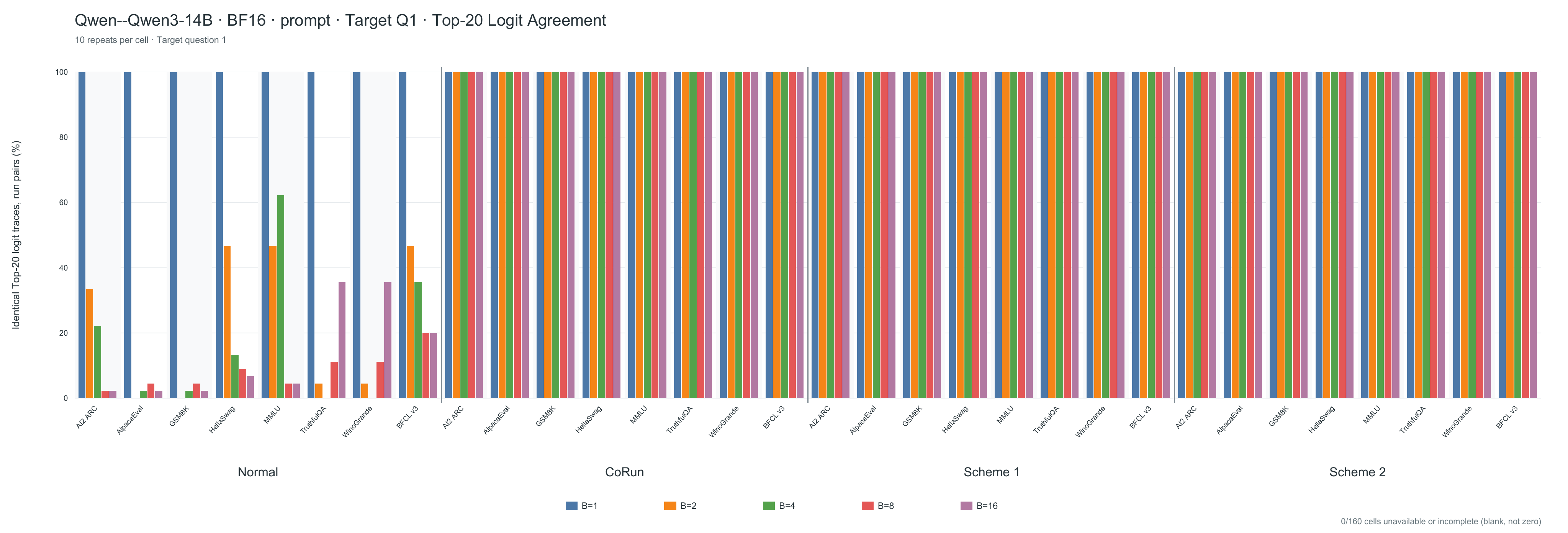}{(b) Top-20 raw-logit agreement.}
\ExpResultPanel{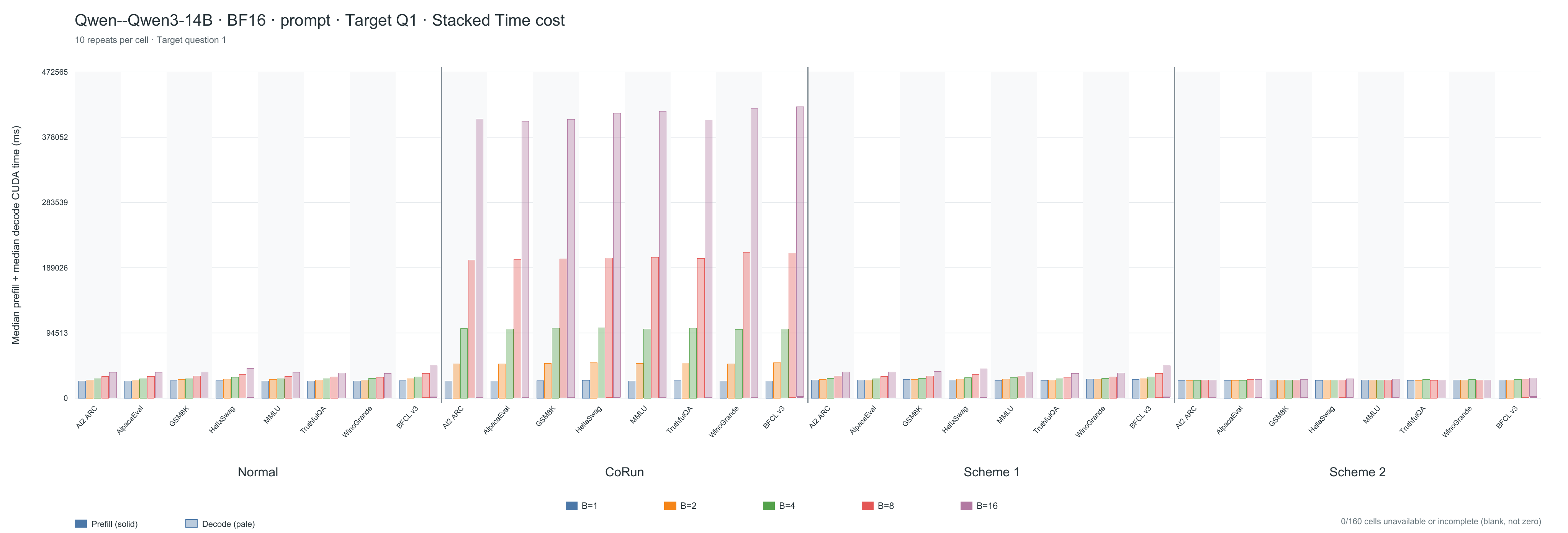}{(c) Prefill and decode CUDA time.}
\caption{Qwen3-14B on NVIDIA RTX A6000 (\texttt{BF16}, \texttt{prompt}, target question Q1, $T=500$): (a) generated-sequence agreement; (b) top-20 raw-logit agreement; (c) prefill and decode cuda time. Method labels follow the correspondence defined at the start of Section~\ref{app:detailed_results}.}
\label{fig:exp-a6000-5-qwen3-14b-prompt}
\end{figure*}

\begin{figure*}[p]
\centering
\ExpResultPanel{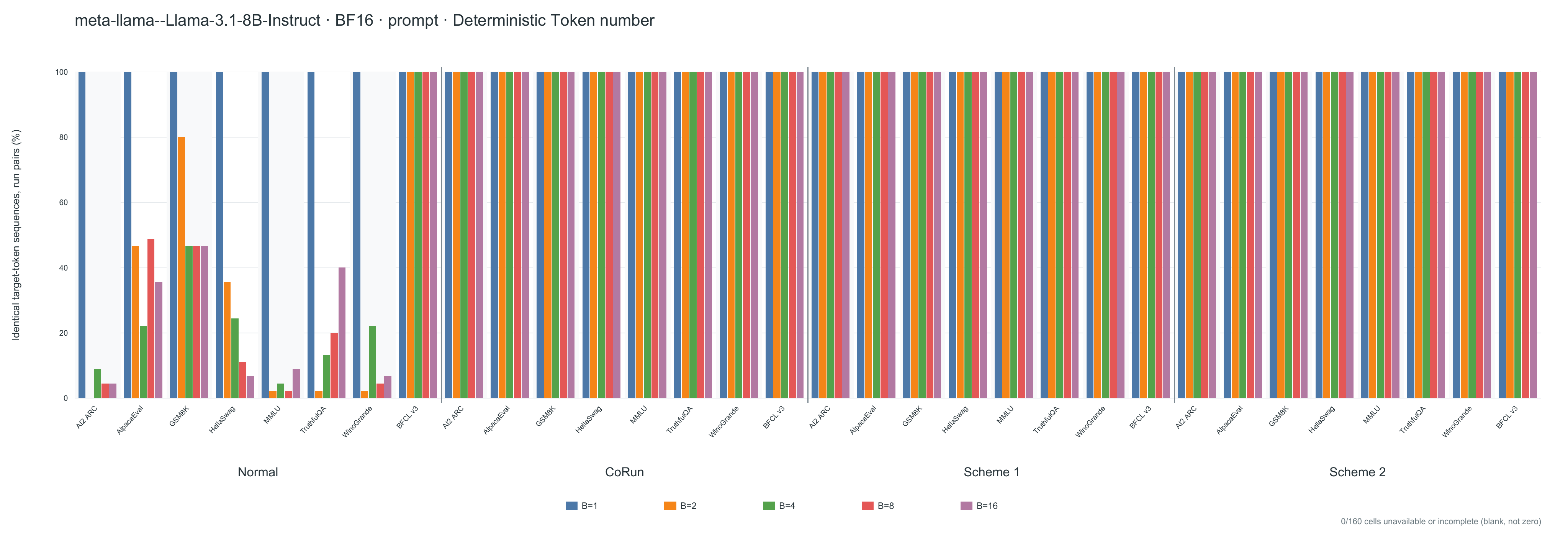}{(a) Generated-sequence agreement.}
\ExpResultPanel{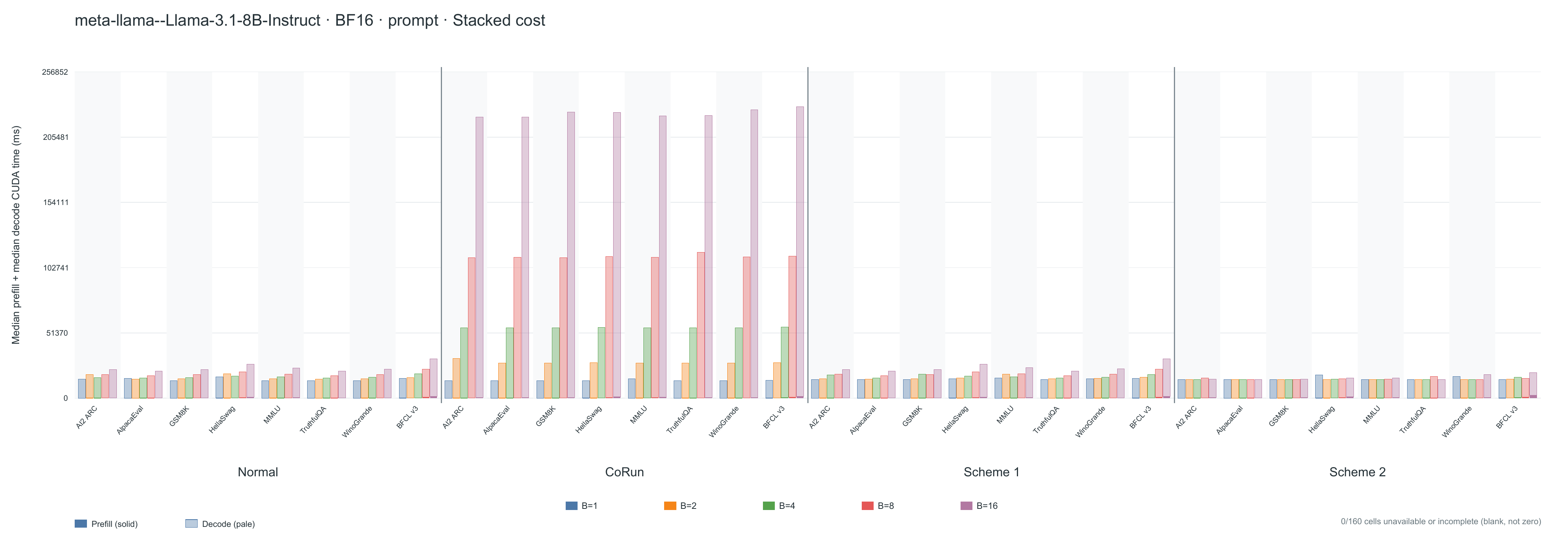}{(b) Prefill and decode CUDA time.}
\caption{Llama-3.1-8B-Instruct on NVIDIA RTX A6000 (\texttt{BF16}, \texttt{prompt}, $T=500$): (a) generated-sequence agreement; (b) prefill and decode cuda time. Method labels follow the correspondence defined at the start of Section~\ref{app:detailed_results}.}
\label{fig:exp-a6000-5-llama-3-1-8b-instruct-prompt}
\end{figure*}

\begin{figure*}[p]
\centering
\ExpResultPanel{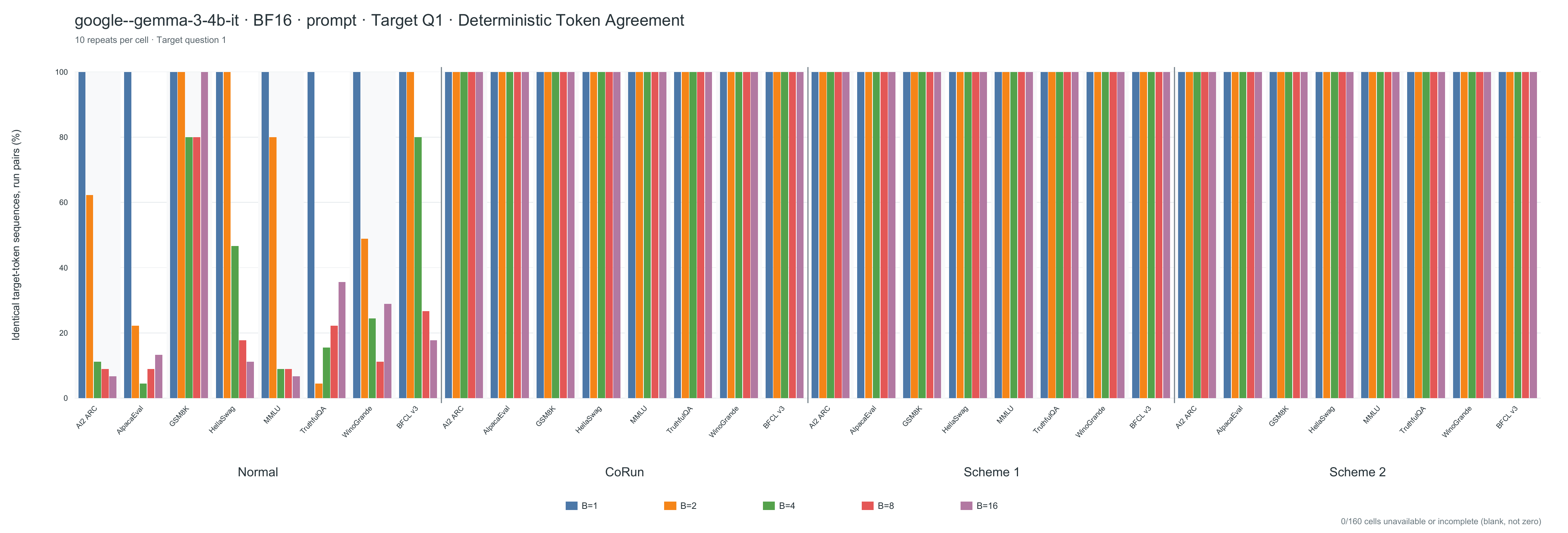}{(a) Generated-sequence agreement.}
\ExpResultPanel{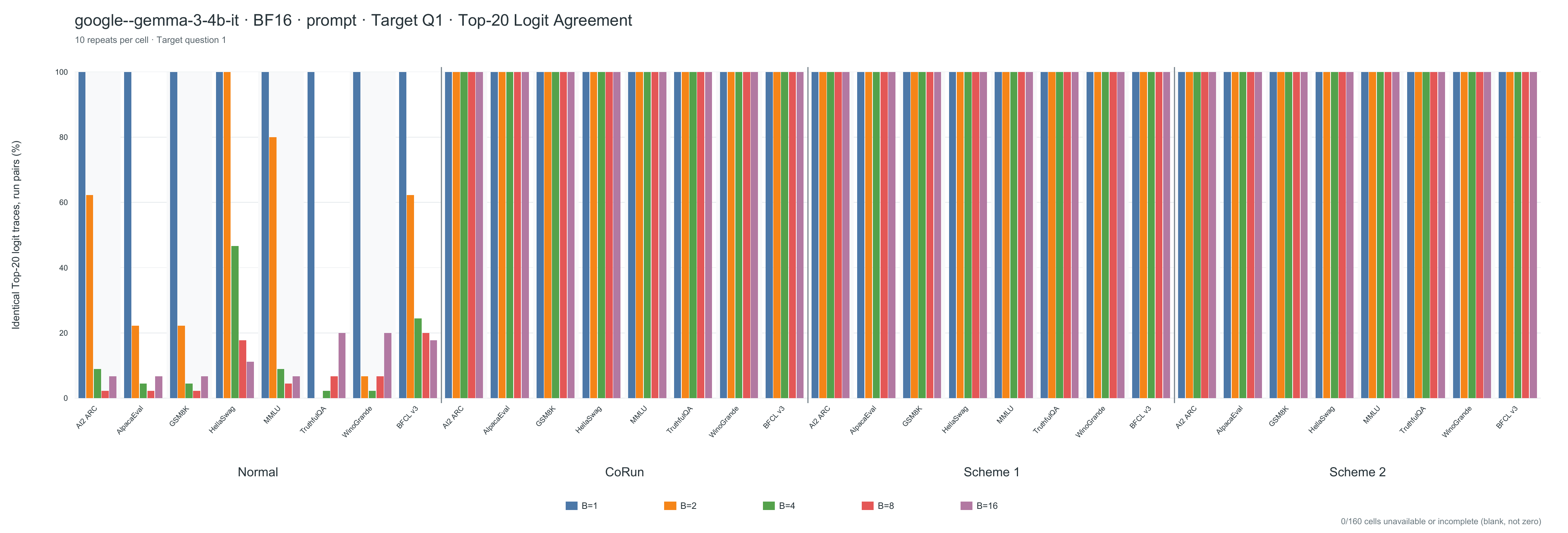}{(b) Top-20 raw-logit agreement.}
\ExpResultPanel{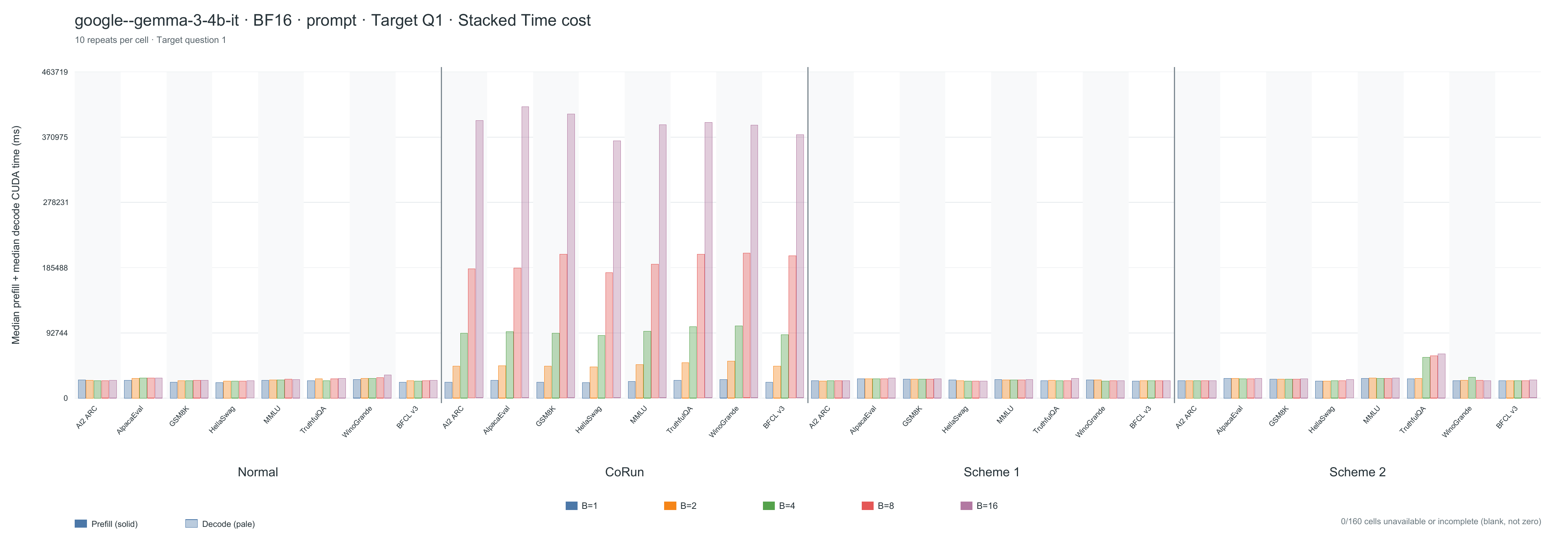}{(c) Prefill and decode CUDA time.}
\caption{Gemma-3-4B-IT on NVIDIA RTX A6000 (\texttt{BF16}, \texttt{prompt}, target question Q1, $T=500$): (a) generated-sequence agreement; (b) top-20 raw-logit agreement; (c) prefill and decode cuda time. Method labels follow the correspondence defined at the start of Section~\ref{app:detailed_results}.}
\label{fig:exp-a6000-5-gemma-3-4b-it-prompt}
\end{figure*}

\begin{figure*}[p]
\centering
\ExpResultPanel{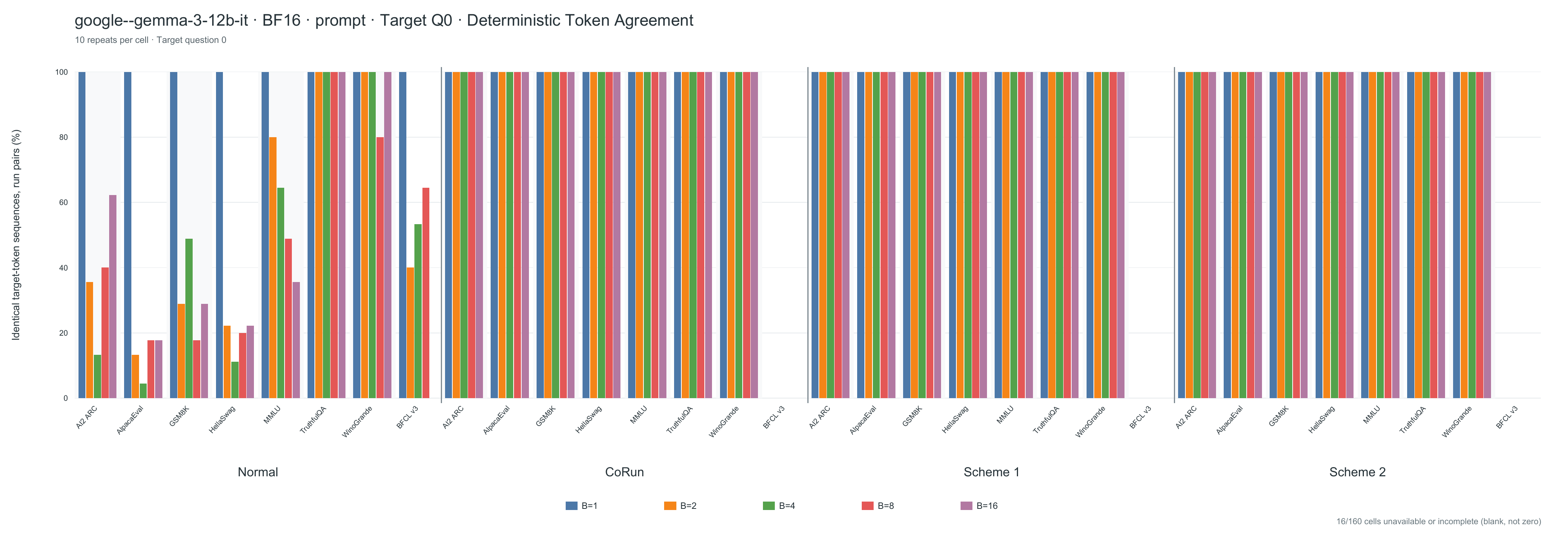}{(a) Generated-sequence agreement.}
\ExpResultPanel{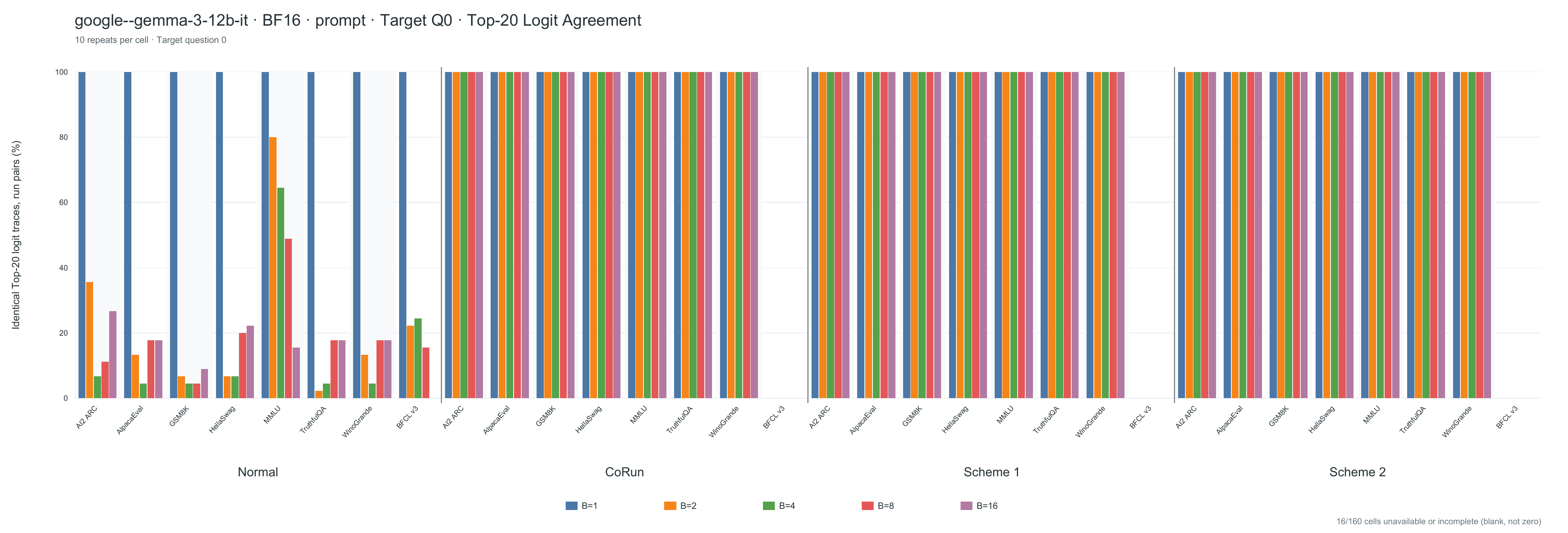}{(b) Top-20 raw-logit agreement.}
\ExpResultPanel{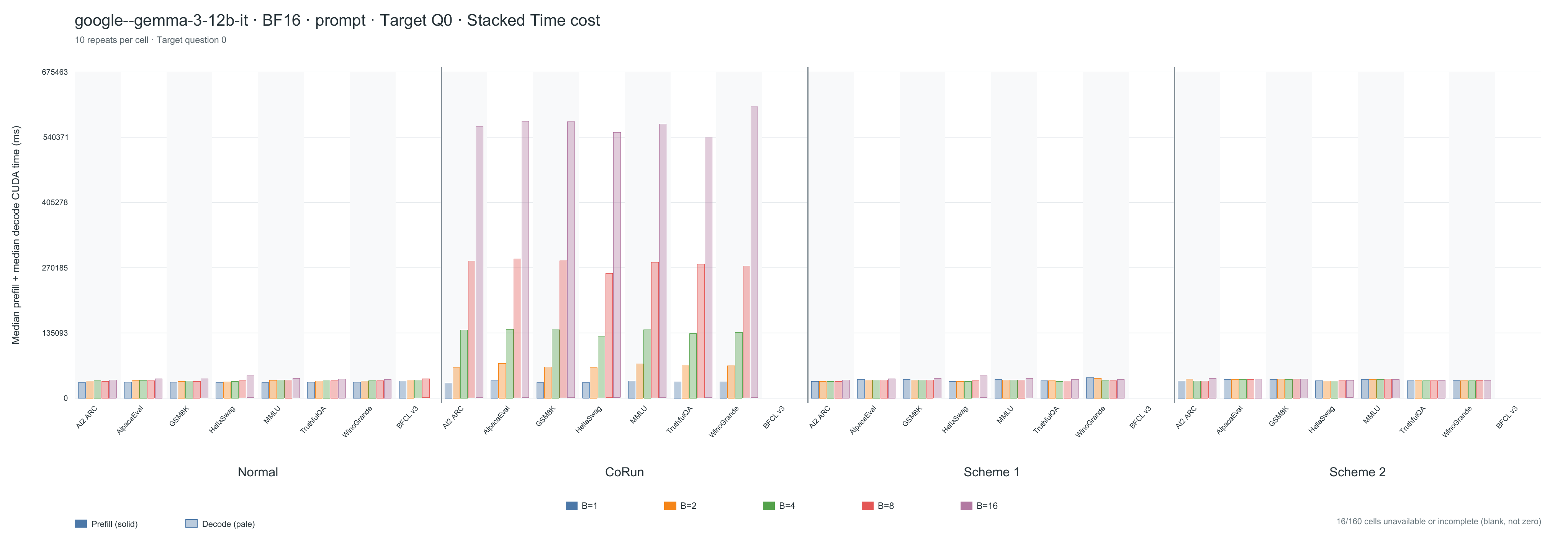}{(c) Prefill and decode CUDA time.}
\caption{Gemma-3-12B-IT on NVIDIA RTX A6000 (\texttt{BF16}, \texttt{prompt}, target question Q0, $T=500$): (a) generated-sequence agreement; (b) top-20 raw-logit agreement; (c) prefill and decode cuda time. Method labels follow the correspondence defined at the start of Section~\ref{app:detailed_results}.}
\label{fig:exp-a6000-5-gemma-3-12b-it-prompt}
\end{figure*}

\clearpage

\begin{figure*}[p]
\subsection{NVIDIA A100}
\label{app:results-a100}

\centering
\ExpResultPanel{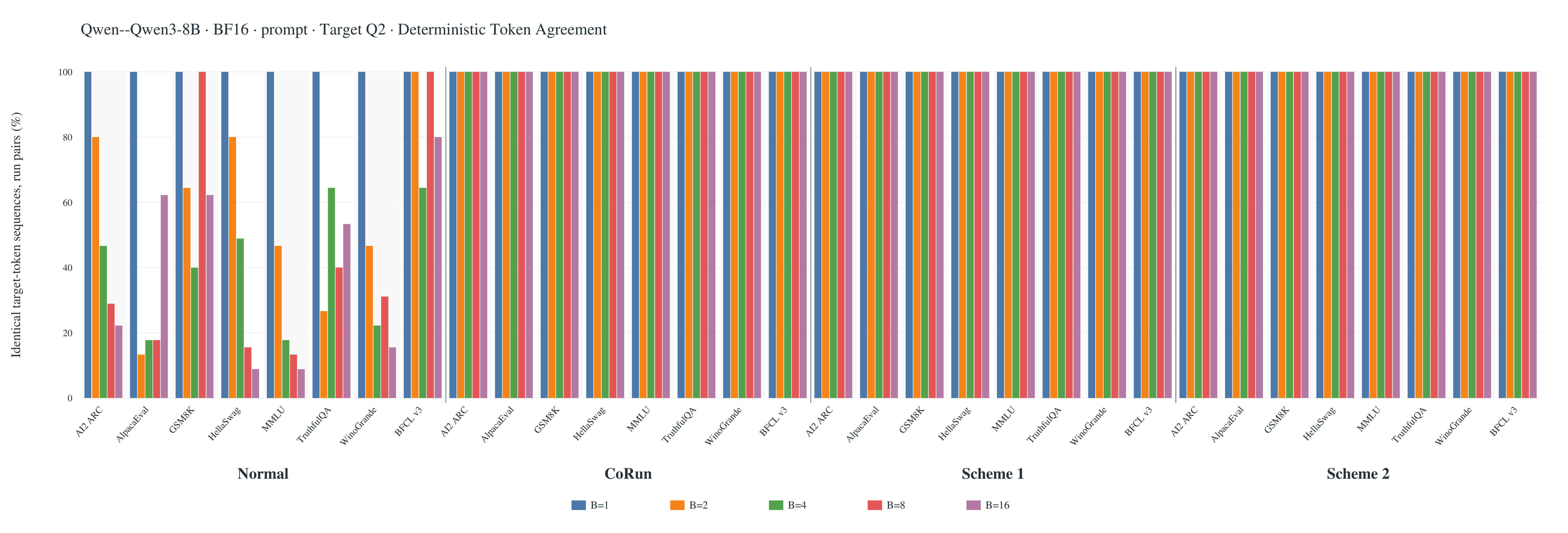}{(a) Generated-sequence agreement.}
\ExpResultPanel{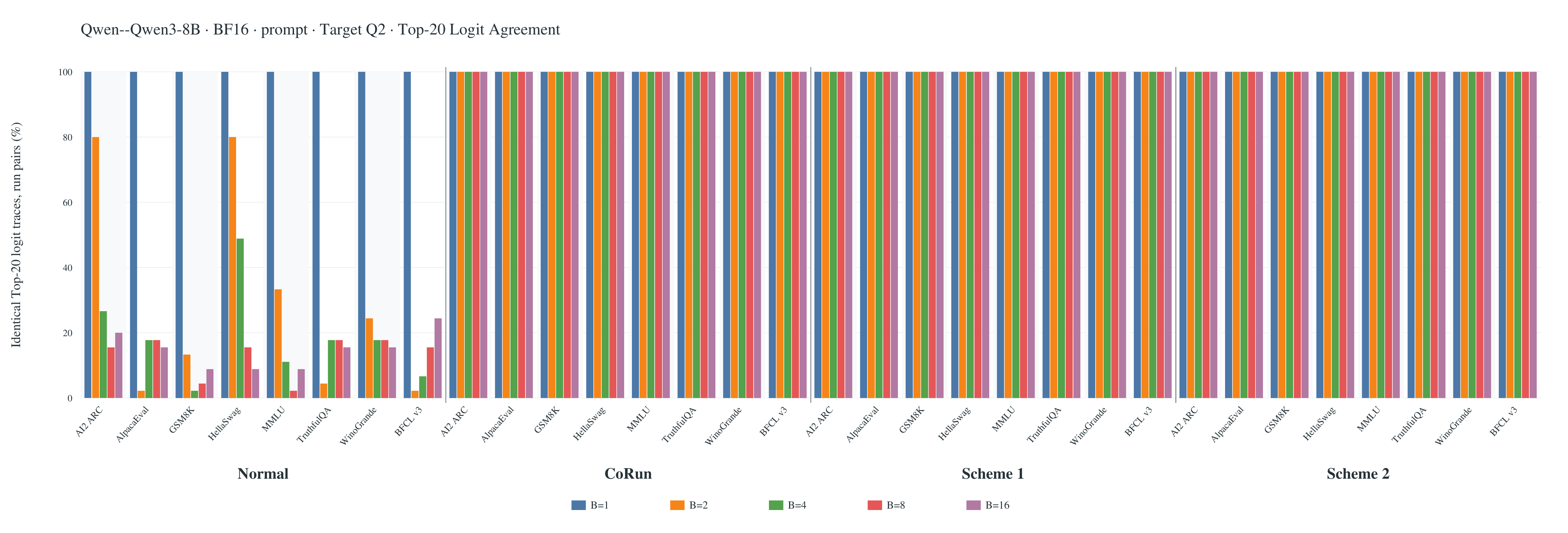}{(b) Top-20 raw-logit agreement.}
\ExpResultPanel{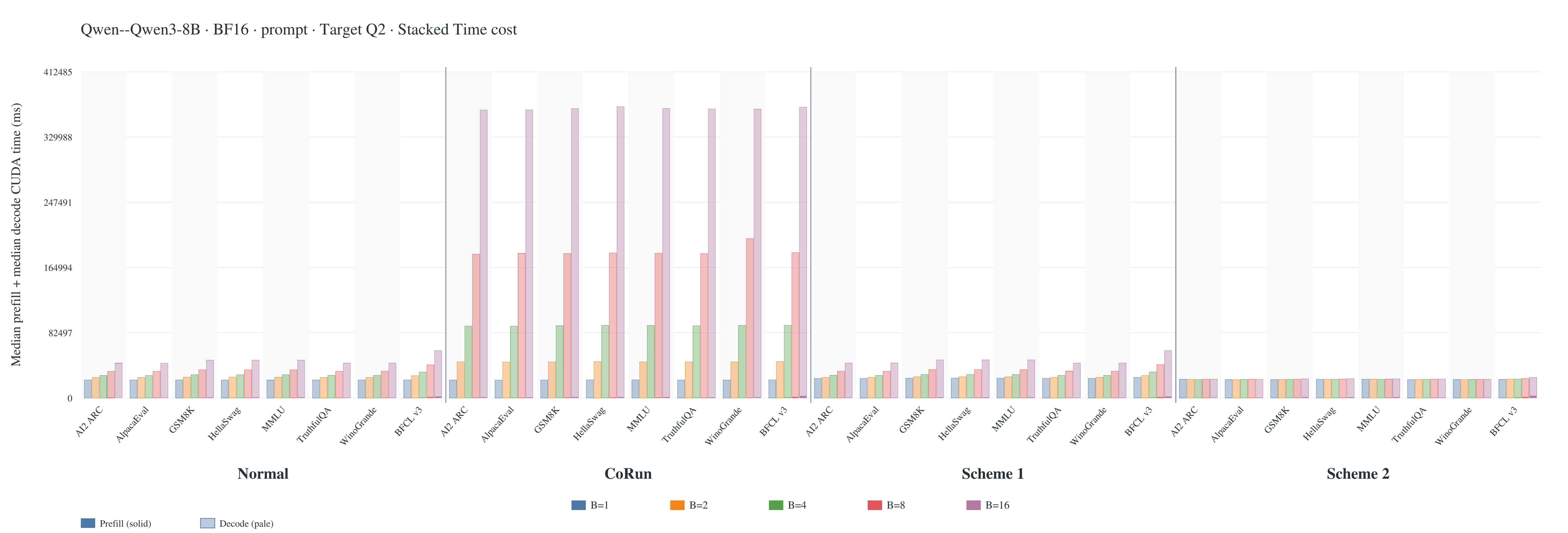}{(c) Prefill and decode CUDA time.}
\caption{Qwen3-8B on NVIDIA A100 (\texttt{BF16}, \texttt{prompt}, target question Q2, $T=500$): (a) generated-sequence agreement; (b) top-20 raw-logit agreement; (c) prefill and decode cuda time. Method labels follow the correspondence defined at the start of Section~\ref{app:detailed_results}.}
\label{fig:exp-a100-qwen3-8b-prompt}
\end{figure*}

\begin{figure*}[p]
\centering
\ExpResultPanel{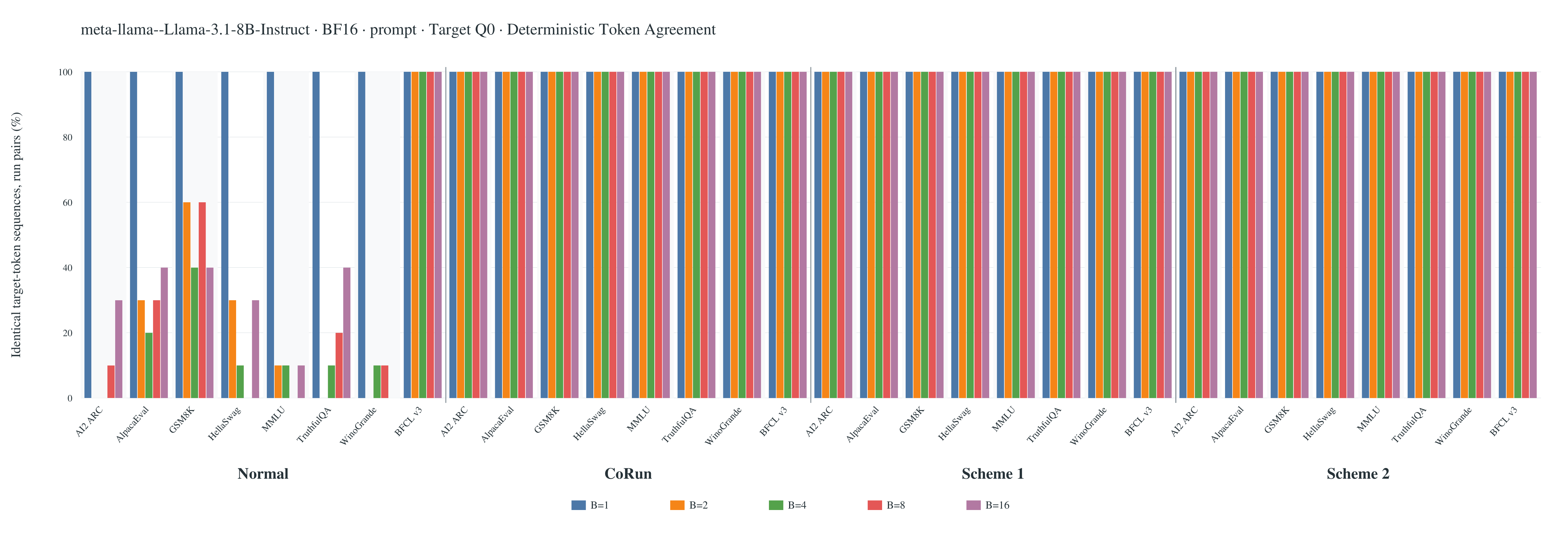}{(a) Generated-sequence agreement.}
\ExpResultPanel{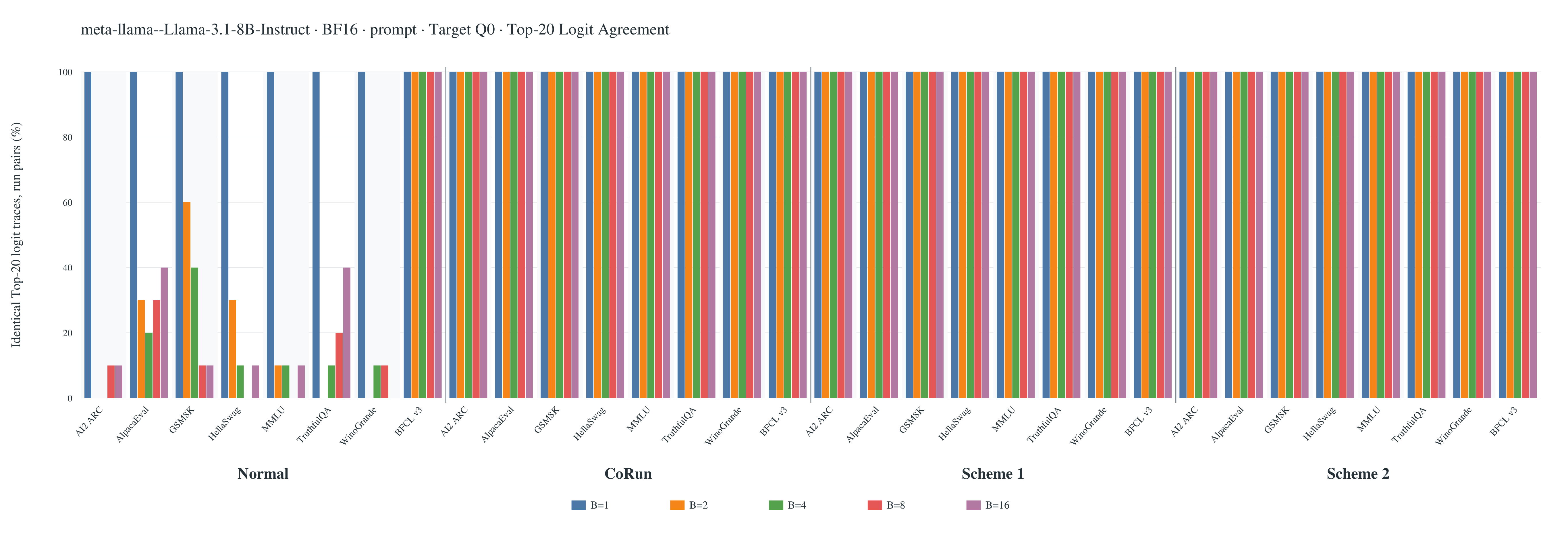}{(b) Top-20 raw-logit agreement.}
\ExpResultPanel{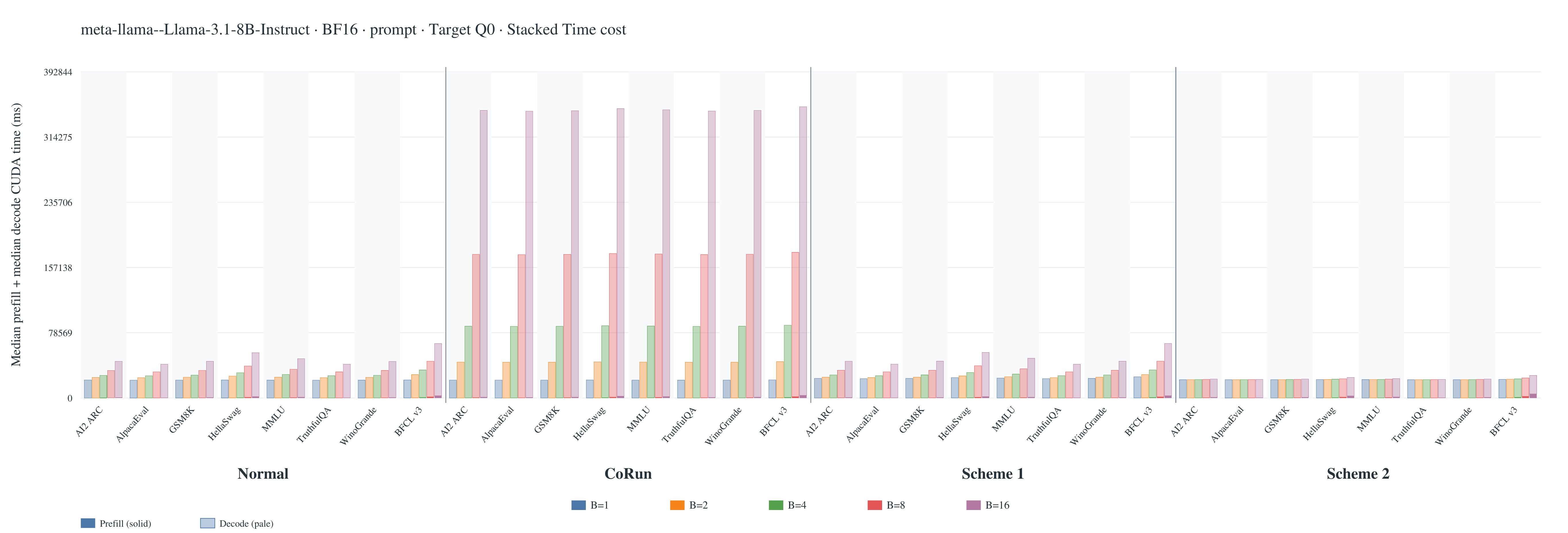}{(c) Prefill and decode CUDA time.}
\caption{Llama-3.1-8B-Instruct on NVIDIA A100 (\texttt{BF16}, \texttt{prompt}, target question Q0, $T=500$): (a) generated-sequence agreement; (b) top-20 raw-logit agreement; (c) prefill and decode cuda time. Method labels follow the correspondence defined at the start of Section~\ref{app:detailed_results}.}
\label{fig:exp-a100-llama-3-1-8b-instruct-prompt}
\end{figure*}

\begin{figure*}[p]
\centering
\ExpResultPanel{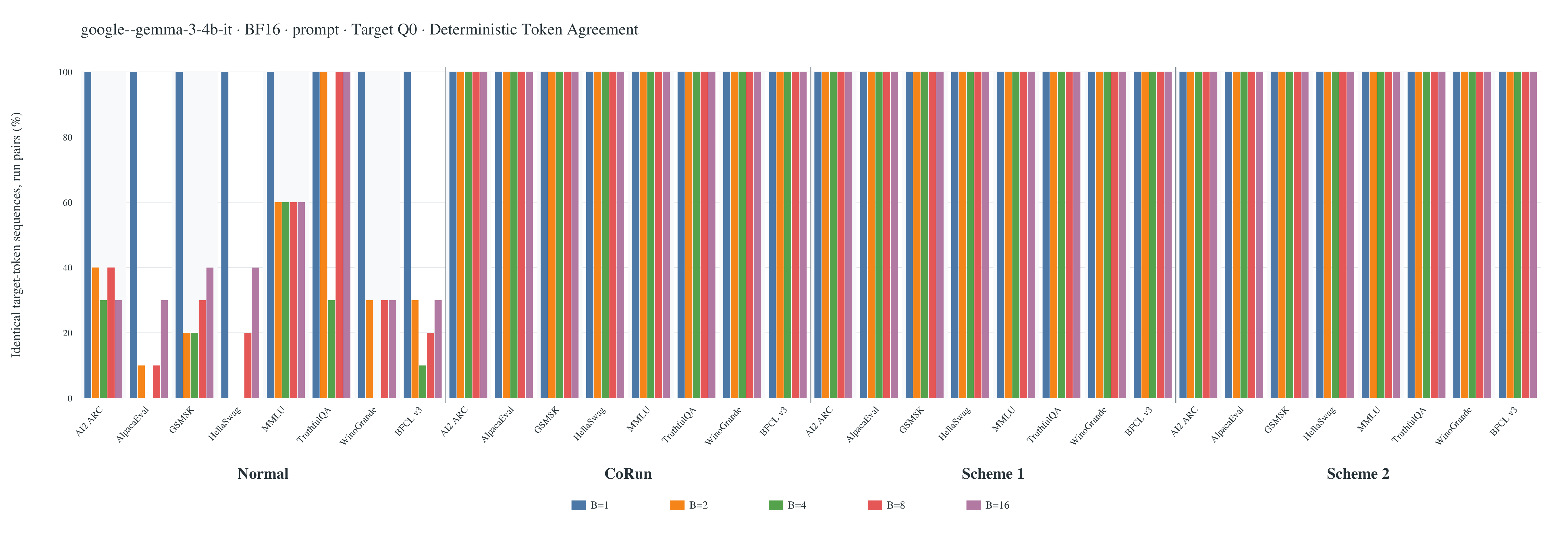}{(a) Generated-sequence agreement.}
\ExpResultPanel{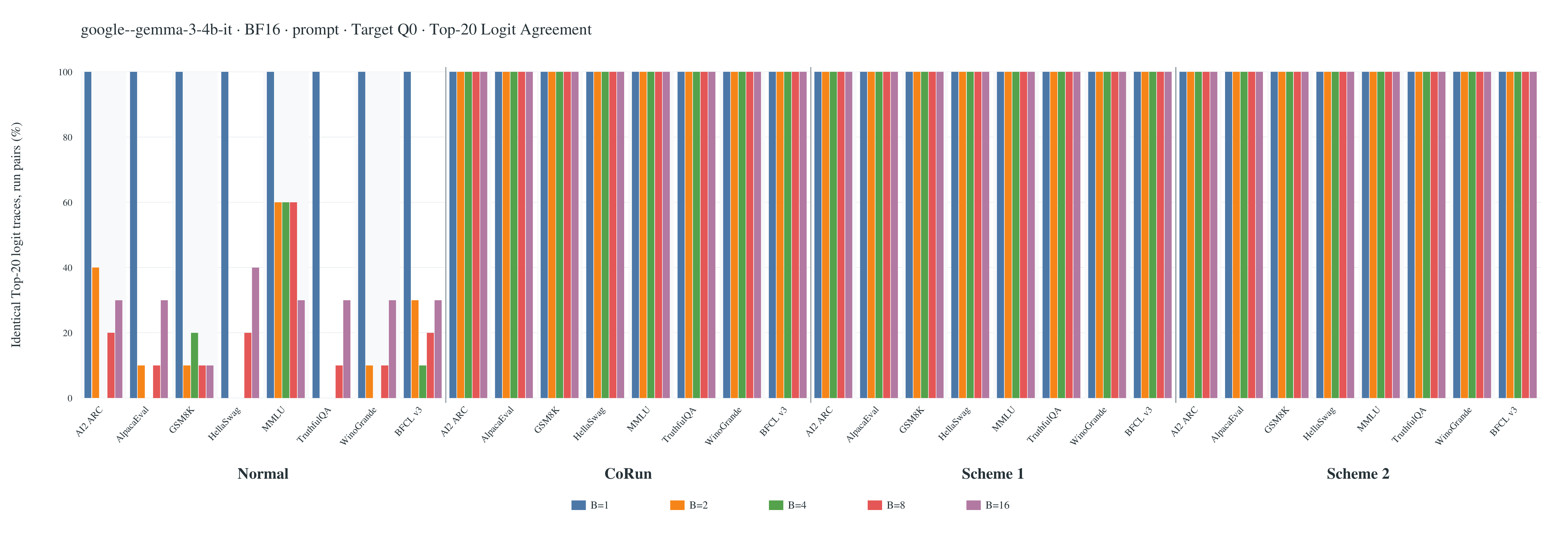}{(b) Top-20 raw-logit agreement.}
\ExpResultPanel{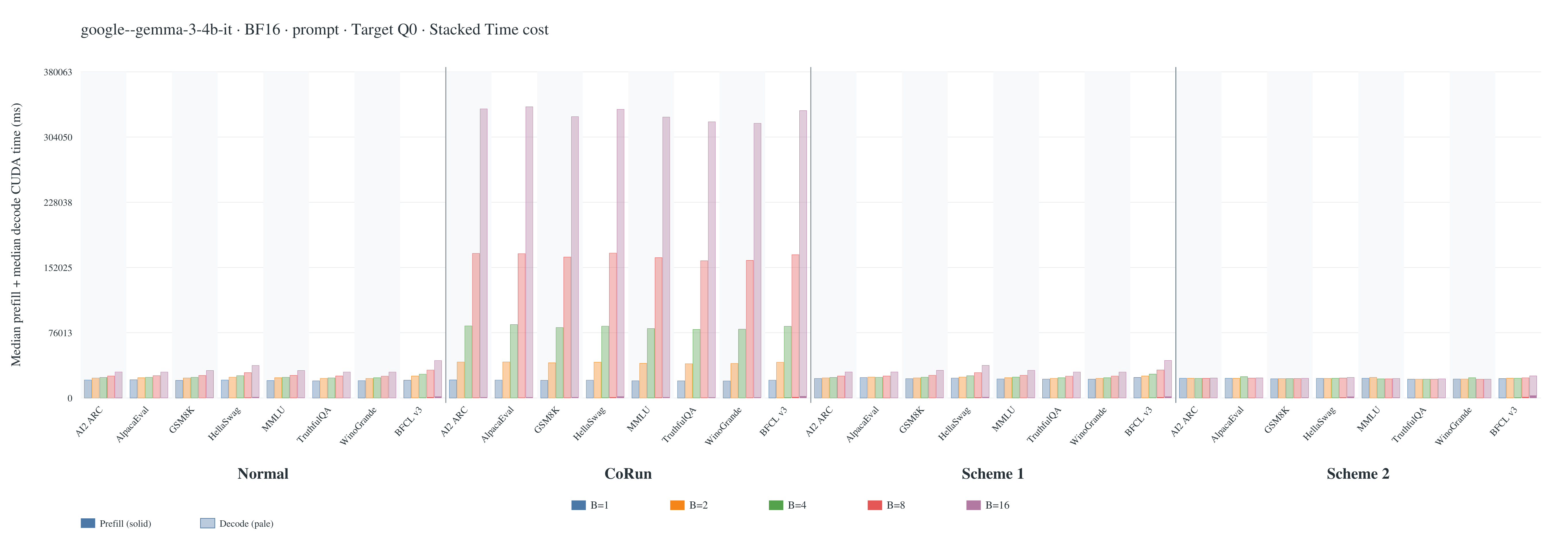}{(c) Prefill and decode CUDA time.}
\caption{Gemma-3-4B-IT on NVIDIA A100 (\texttt{BF16}, \texttt{prompt}, target question Q0, $T=500$): (a) generated-sequence agreement; (b) top-20 raw-logit agreement; (c) prefill and decode cuda time. Method labels follow the correspondence defined at the start of Section~\ref{app:detailed_results}.}
\label{fig:exp-a100-gemma-3-4b-it-prompt}
\end{figure*}

\begin{figure*}[p]
\centering
\ExpResultPanel{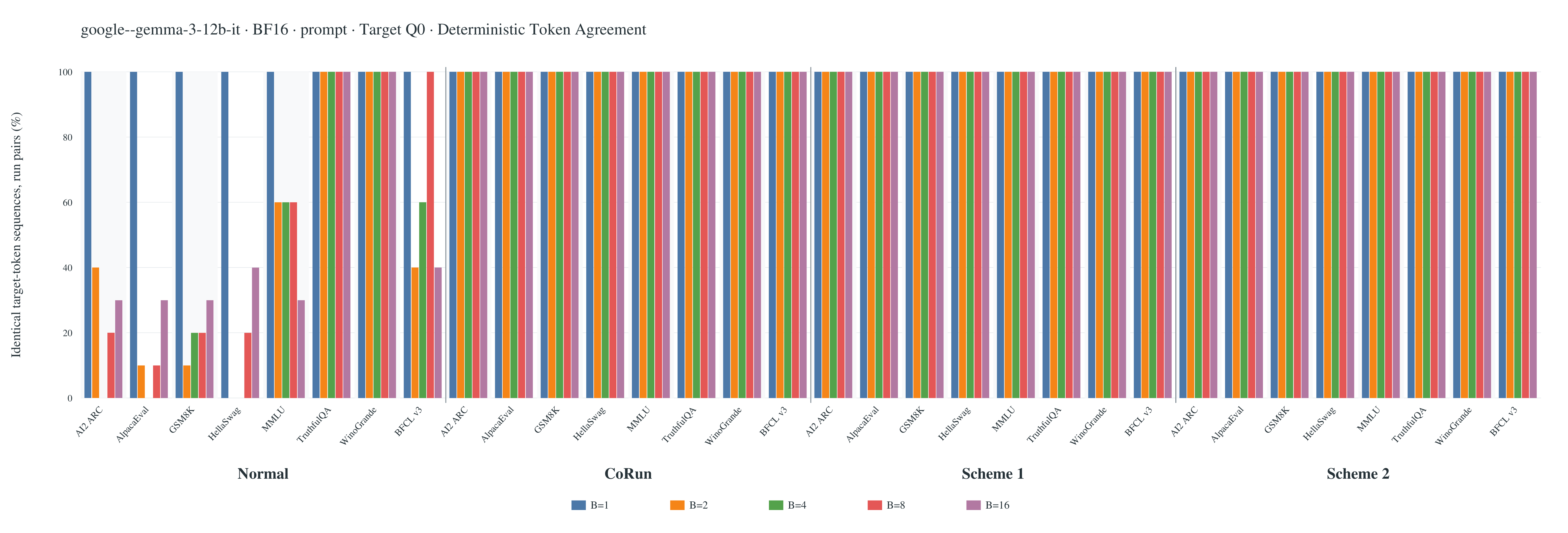}{(a) Generated-sequence agreement.}
\ExpResultPanel{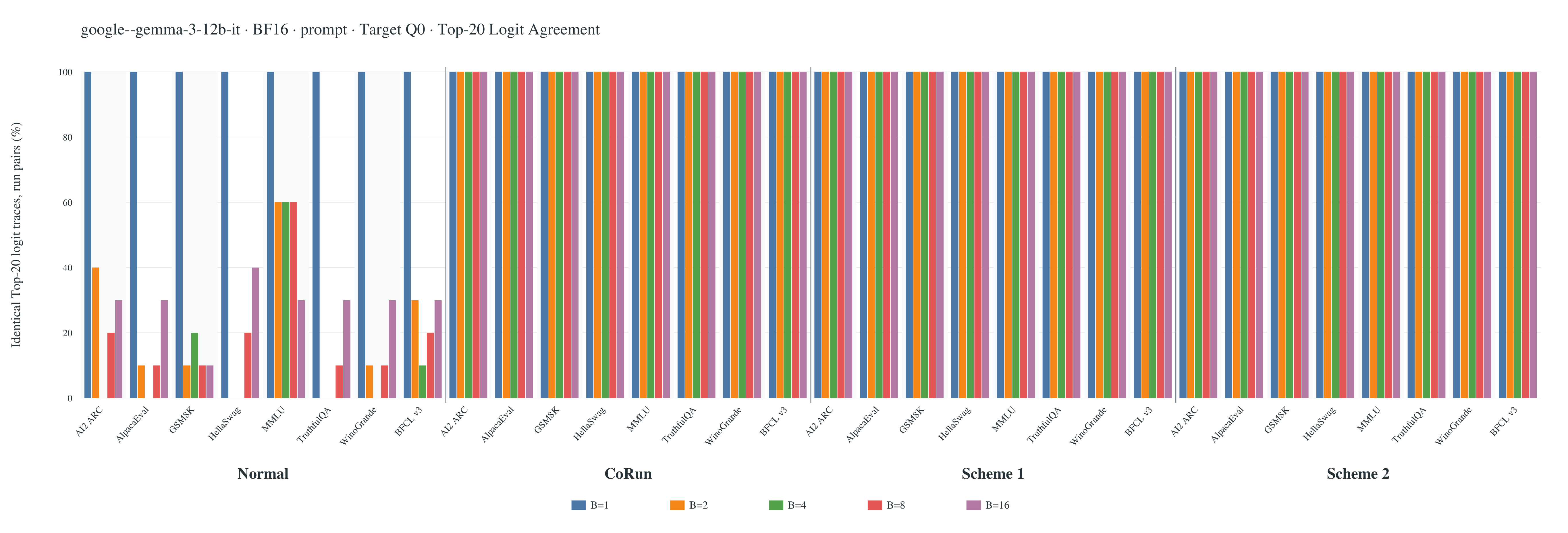}{(b) Top-20 raw-logit agreement.}
\ExpResultPanel{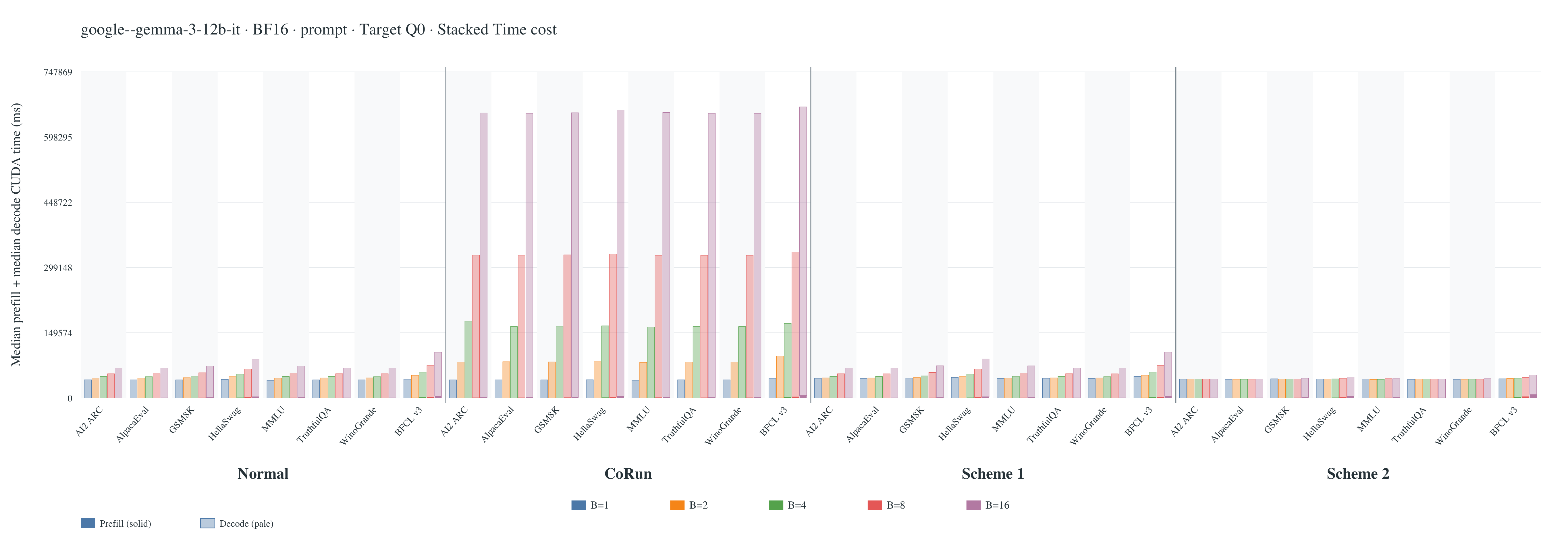}{(c) Prefill and decode CUDA time.}
\caption{Gemma-3-12B-IT on NVIDIA A100 (\texttt{BF16}, \texttt{prompt}, target question Q0, $T=500$): (a) generated-sequence agreement; (b) top-20 raw-logit agreement; (c) prefill and decode cuda time. Method labels follow the correspondence defined at the start of Section~\ref{app:detailed_results}.}
\label{fig:exp-a100-gemma-3-12b-it-prompt}
\end{figure*}

\clearpage

\begin{figure*}[p]
\subsection{NVIDIA H100}
\label{app:results-h100-5}

\centering
\ExpResultPanel{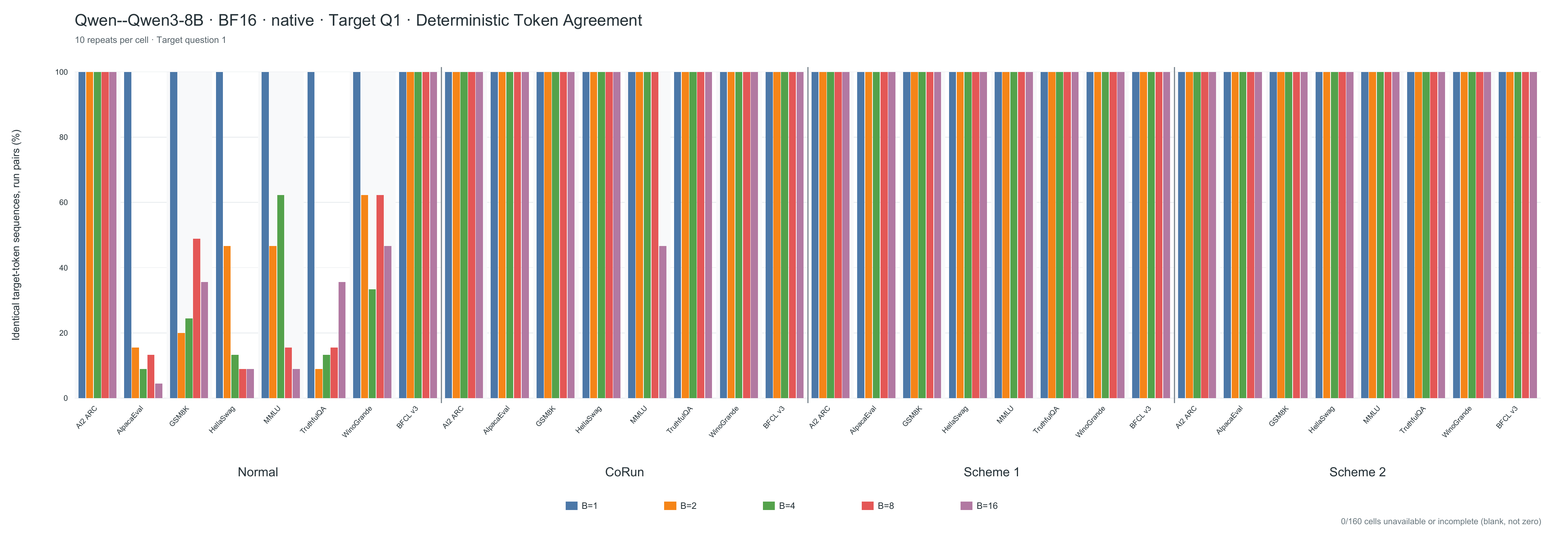}{(a) Generated-sequence agreement.}
\ExpResultPanel{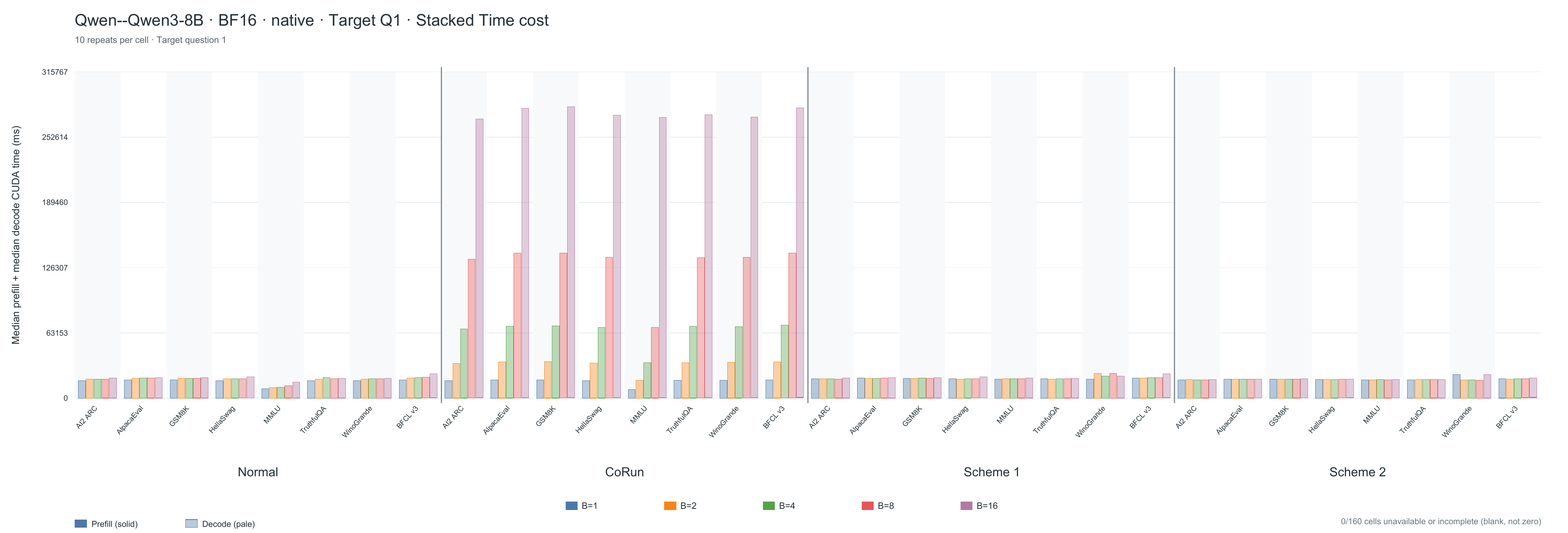}{(b) Prefill and decode CUDA time.}
\caption{Qwen3-8B on NVIDIA H100 (\texttt{BF16}, \texttt{native}, target question Q1, $T=500$): (a) generated-sequence agreement; (b) prefill and decode cuda time. Method labels follow the correspondence defined at the start of Section~\ref{app:detailed_results}.}
\label{fig:exp-h100-5-qwen3-8b-native}
\end{figure*}

\begin{figure*}[p]
\centering
\ExpResultPanel{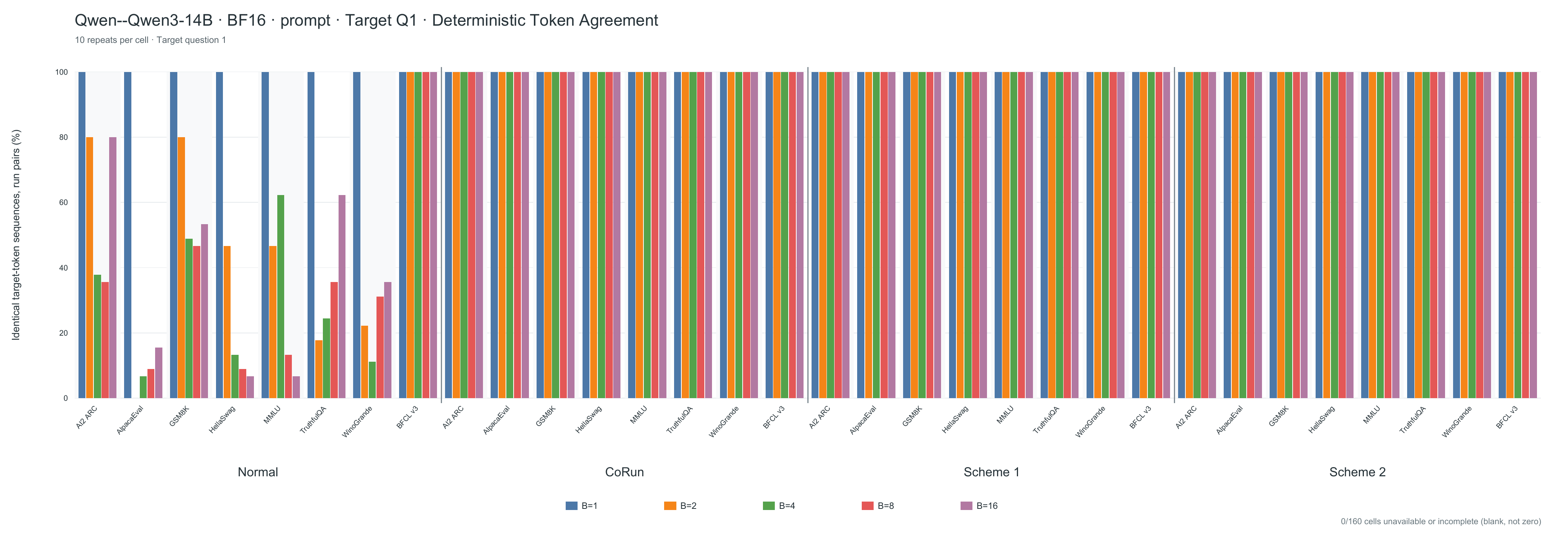}{(a) Generated-sequence agreement.}
\ExpResultPanel{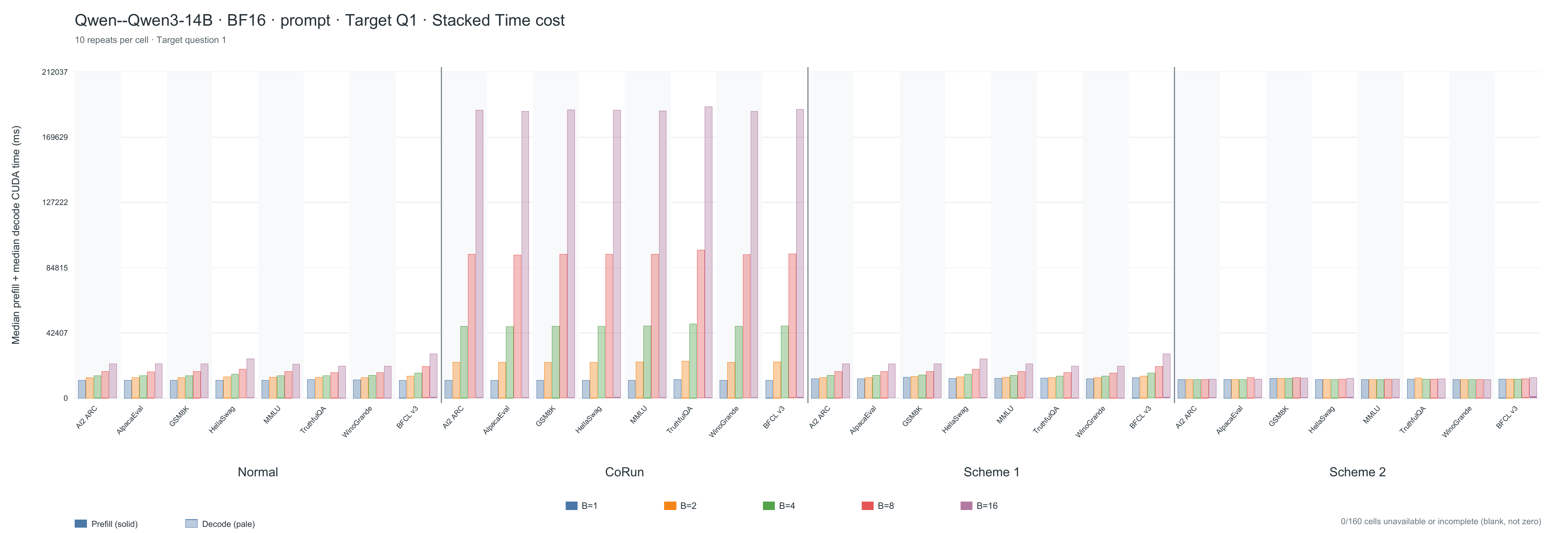}{(b) Prefill and decode CUDA time.}
\caption{Qwen3-14B on NVIDIA H100 (\texttt{BF16}, \texttt{prompt}, target question Q1, $T=500$): (a) generated-sequence agreement; (b) prefill and decode cuda time. Method labels follow the correspondence defined at the start of Section~\ref{app:detailed_results}.}
\label{fig:exp-h100-5-qwen3-14b-prompt}
\end{figure*}

\begin{figure*}[p]
\centering
\ExpResultPanel{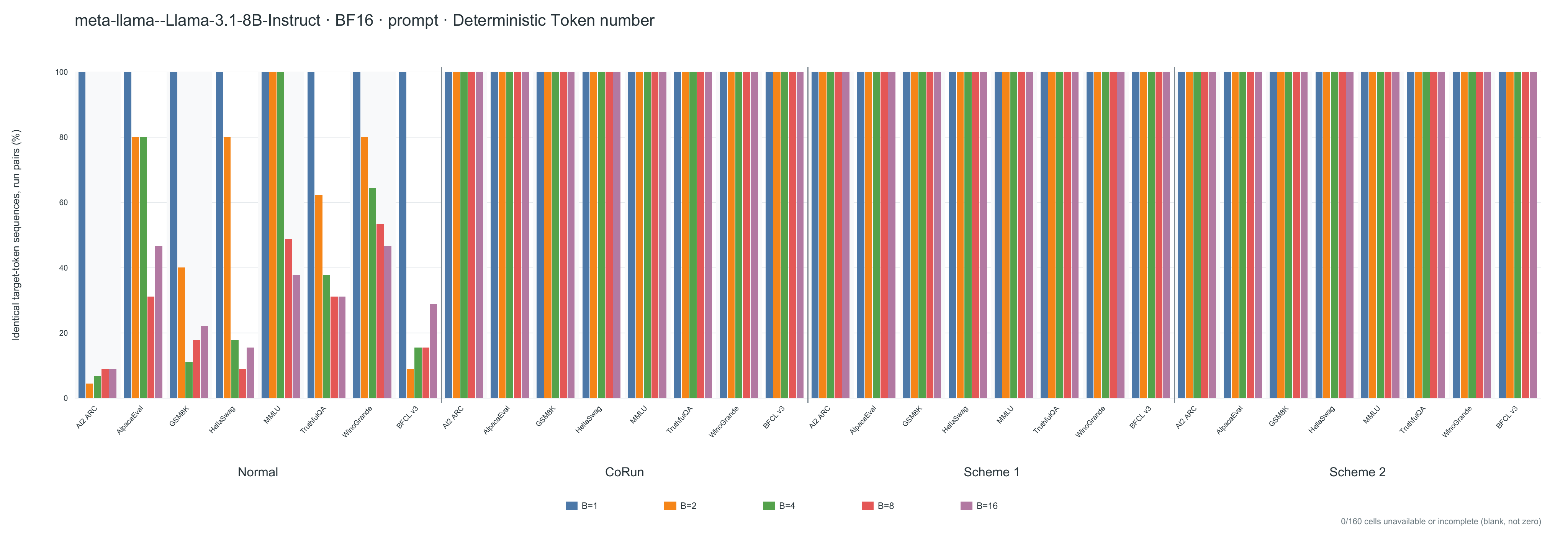}{(a) Generated-sequence agreement.}
\ExpResultPanel{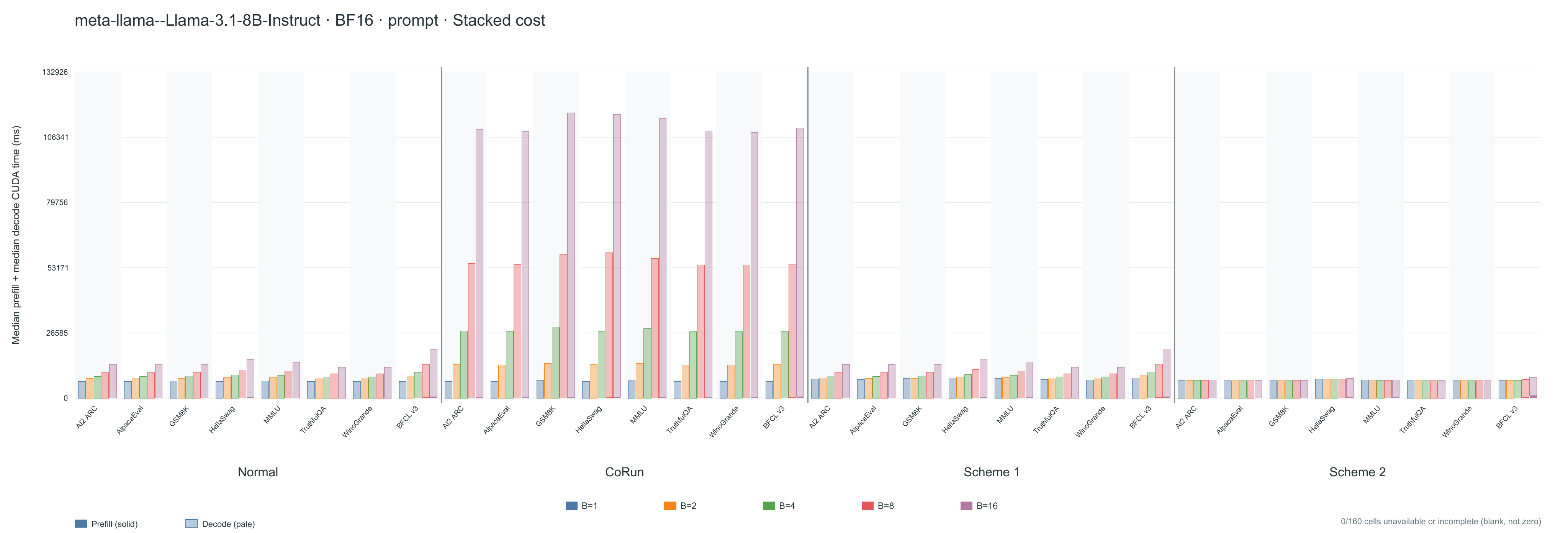}{(b) Prefill and decode CUDA time.}
\caption{Llama-3.1-8B-Instruct on NVIDIA H100 (\texttt{BF16}, \texttt{prompt}, $T=500$): (a) generated-sequence agreement; (b) prefill and decode cuda time. Method labels follow the correspondence defined at the start of Section~\ref{app:detailed_results}.}
\label{fig:exp-h100-5-llama-3-1-8b-instruct-prompt}
\end{figure*}

\begin{figure*}[p]
\centering
\ExpResultPanel{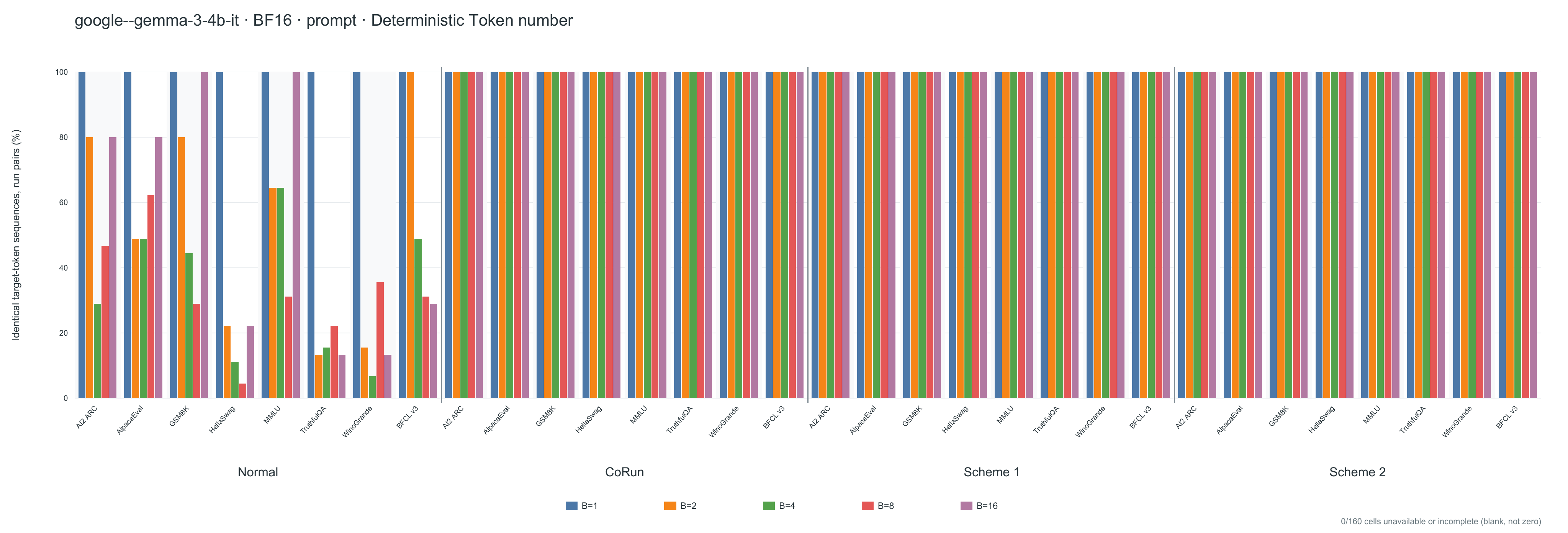}{(a) Generated-sequence agreement.}
\ExpResultPanel{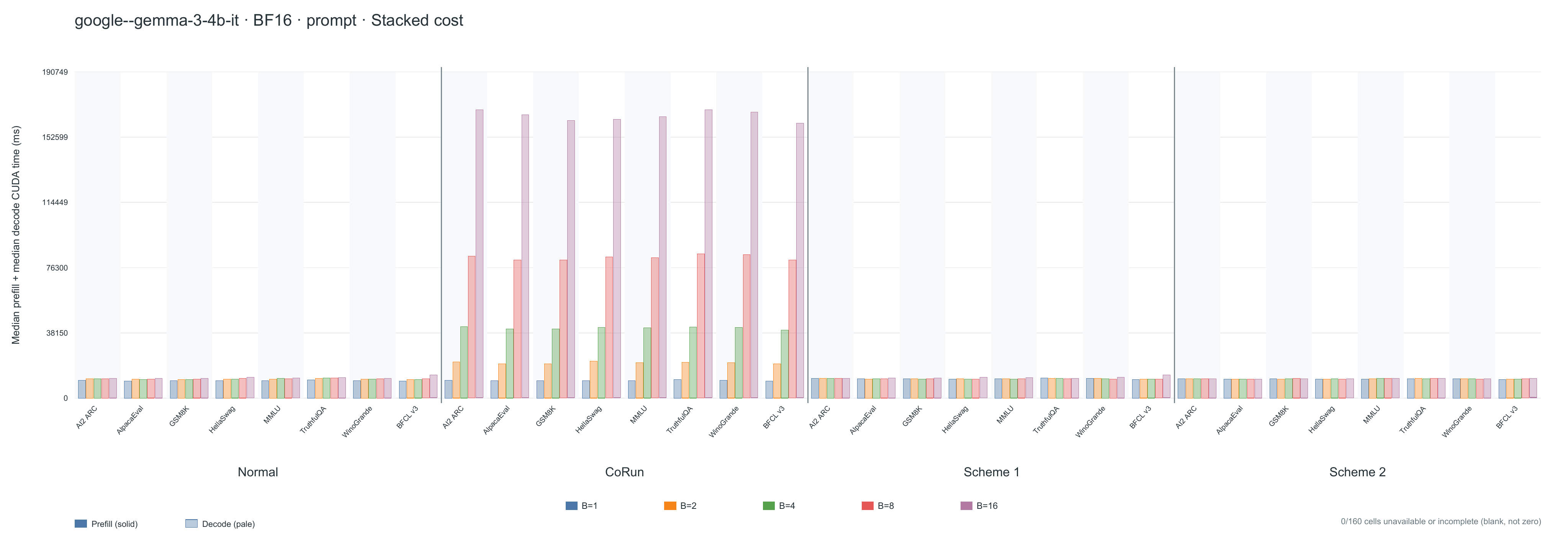}{(b) Prefill and decode CUDA time.}
\caption{Gemma-3-4B-IT on NVIDIA H100 (\texttt{BF16}, \texttt{prompt}, $T=500$): (a) generated-sequence agreement; (b) prefill and decode cuda time. Method labels follow the correspondence defined at the start of Section~\ref{app:detailed_results}.}
\label{fig:exp-h100-5-gemma-3-4b-it-prompt}
\end{figure*}


\clearpage

\clearpage
\bibliography{refs}
\bibliographystyle{mlsys2025}